\documentclass[11pt]{article}

\usepackage[final]{acl}

\usepackage{times}
\usepackage{latexsym}

\usepackage[T1]{fontenc}

\usepackage[utf8]{inputenc}

\usepackage{microtype}

\usepackage{inconsolata}

\usepackage{graphicx}
\usepackage{multirow}
\usepackage{booktabs}
\usepackage{makecell}
\usepackage{natbib}
\usepackage{xspace}
\usepackage[table]{xcolor}
\usepackage{enumitem}
\setitemize[1]{itemsep=-1pt,partopsep=0pt,topsep=1pt}
\usepackage{amsmath}
\usepackage{mathtools}
\usepackage{amssymb}
\usepackage{tcolorbox}
\usepackage[dvipsnames]{xcolor}
\usepackage{pifont}
\usepackage{url}

\newcommand{\icoyes}{\textcolor{ForestGreen}{\ding{51}}\xspace}
\newcommand{\icono}{\textcolor{BrickRed}{\ding{55}}\xspace}

\title{\raisebox{-0.3\height}{\includegraphics[height=1.6em]{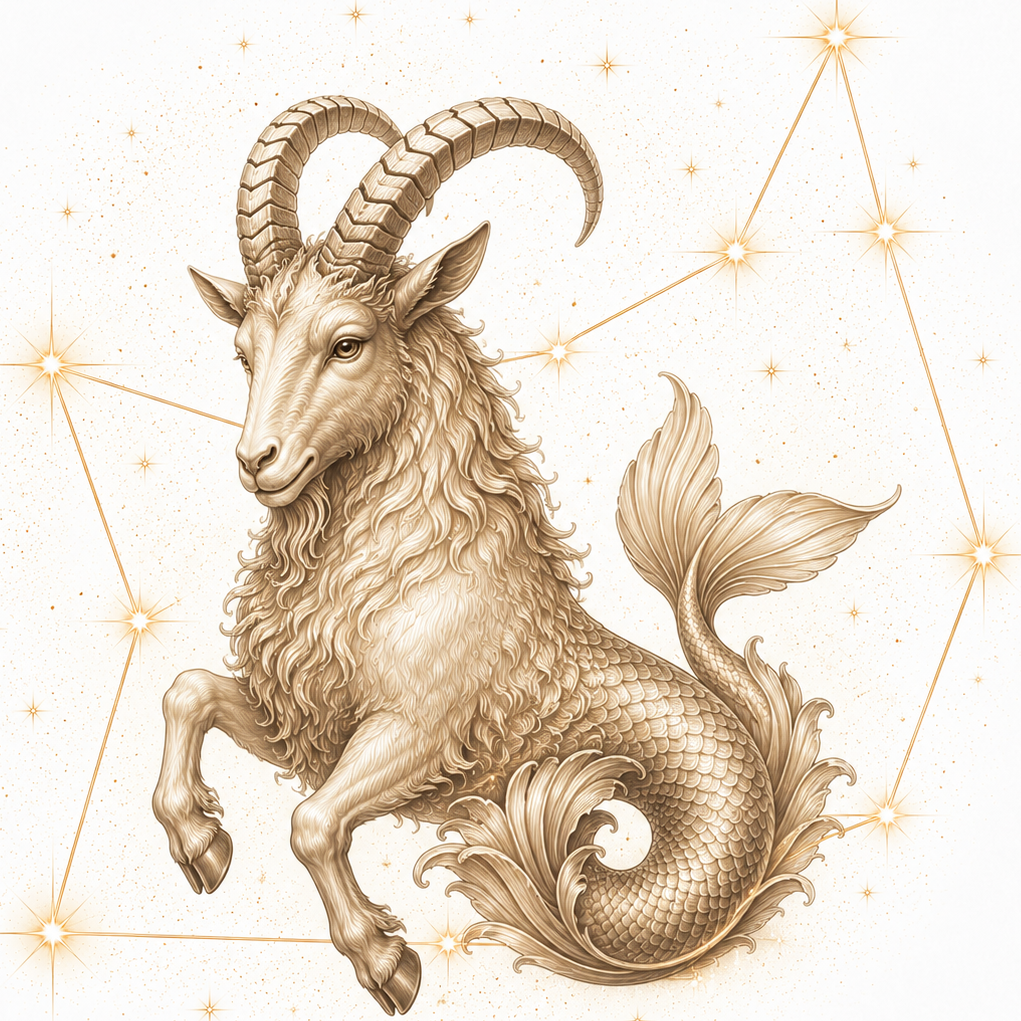}} CapRiCorn-1K: A Comprehensive Benchmark for Video Captioning and Subject Referential Consistency Across Temporal Scales}

\author{Xinlong Chen\textsuperscript{\textmd{1,2,3}}\thanks{This work was conducted during the author's internship at Kling Team, Kuaishou Technology},
Jiafu Tang\textsuperscript{\textmd{4}},
Yue Ding\textsuperscript{\textmd{1,2}},
Yizhuo Jia\textsuperscript{\textmd{5}},
\textbf{Bozhou Li}\textsuperscript{\textmd{6}},
\textbf{Bohan Zeng}\textsuperscript{\textmd{6}}, \\
\textbf{Yang Shi}\textsuperscript{\textmd{6}},
\textbf{Shihao Li}\textsuperscript{\textmd{4}},
\textbf{Yiyan Ji}\textsuperscript{\textmd{4}},
\textbf{Qiang Liu}\textsuperscript{\textmd{1,2}\thanks{Corresponding author: qiang.liu@nlpr.ia.ac.cn}},
\textbf{Weihong Lin}\textsuperscript{\textmd{3}},
\textbf{Yuanxing Zhang}\textsuperscript{\textmd{3}}, \\
\textbf{Pengfei Wan}\textsuperscript{\textmd{3}},
\textbf{Liang Wang}\textsuperscript{\textmd{1,2}},
\textbf{Tieniu Tan}\textsuperscript{\textmd{1,2,4}} \\
\textsuperscript{\textmd{1}}NLPR, CASIA \quad
\textsuperscript{\textmd{2}}UCAS \quad
\textsuperscript{\textmd{3}}Kling Team \quad
\textsuperscript{\textmd{4}}NJU \quad
\textsuperscript{\textmd{5}}FDU \quad
\textsuperscript{\textmd{6}}PKU \\
}

\begin{document}
\maketitle
\begin{abstract}
Accurate and comprehensive video captions with consistent subject references are critical for downstream understanding and generation tasks. However, few existing benchmarks can objectively and comprehensively evaluate these properties across diverse durations and scenarios, thereby hindering the advancement of video captioning models. To bridge this gap, we propose CapRiCorn-1K, a comprehensive benchmark designed to evaluate both video captioning quality and subject referential consistency across long temporal horizons and diverse video domains. To accommodate varied evaluation needs, our benchmark supports both audiovisual and visual-only settings. Extensive experiments on CapRiCorn-1K reveal that current models generally struggle to generate accurate and comprehensive captions while maintaining consistent subject references. Moreover, as video duration increases, both the overall caption quality and subject referential consistency decline. Notably, our evaluation metrics exhibit strong correlations with the performance of downstream understanding and generation tasks conditioned on the generated captions, further validating their effectiveness. The project is available at \url{https://github.com/xlchen0205/CapRiCorn-1K}.
\end{abstract}

\section{Introduction}
With the rapid advancement of Multimodal Large Language Models (MLLMs), video captioning has evolved from a basic descriptive task into a core semantic interface that bridges multimodal perception with linguistic semantics~\citep{chen2025avocado, tang2025video, li2026towards}. High-quality video captions not only facilitate the effective alignment of audio, visual, and textual modalities during pre-training~\citep{xu2025qwen3, team2025longcat}, but also inject crucial semantic knowledge into downstream multimodal understanding and generation tasks~\citep{long2025seeing, du2025vc4vg, shi2025mavors, hua2026vabench}. Extensive research has demonstrated that enhancing the quality of video captions yields stable and significant performance gains across a wide range of applications~\citep{team2026script, chen2024sharegpt4video, wang2025haic, an2025onestory, ding2026omnisift}.

\begin{figure*}[t]
    \centering
    \includegraphics[width=\linewidth]{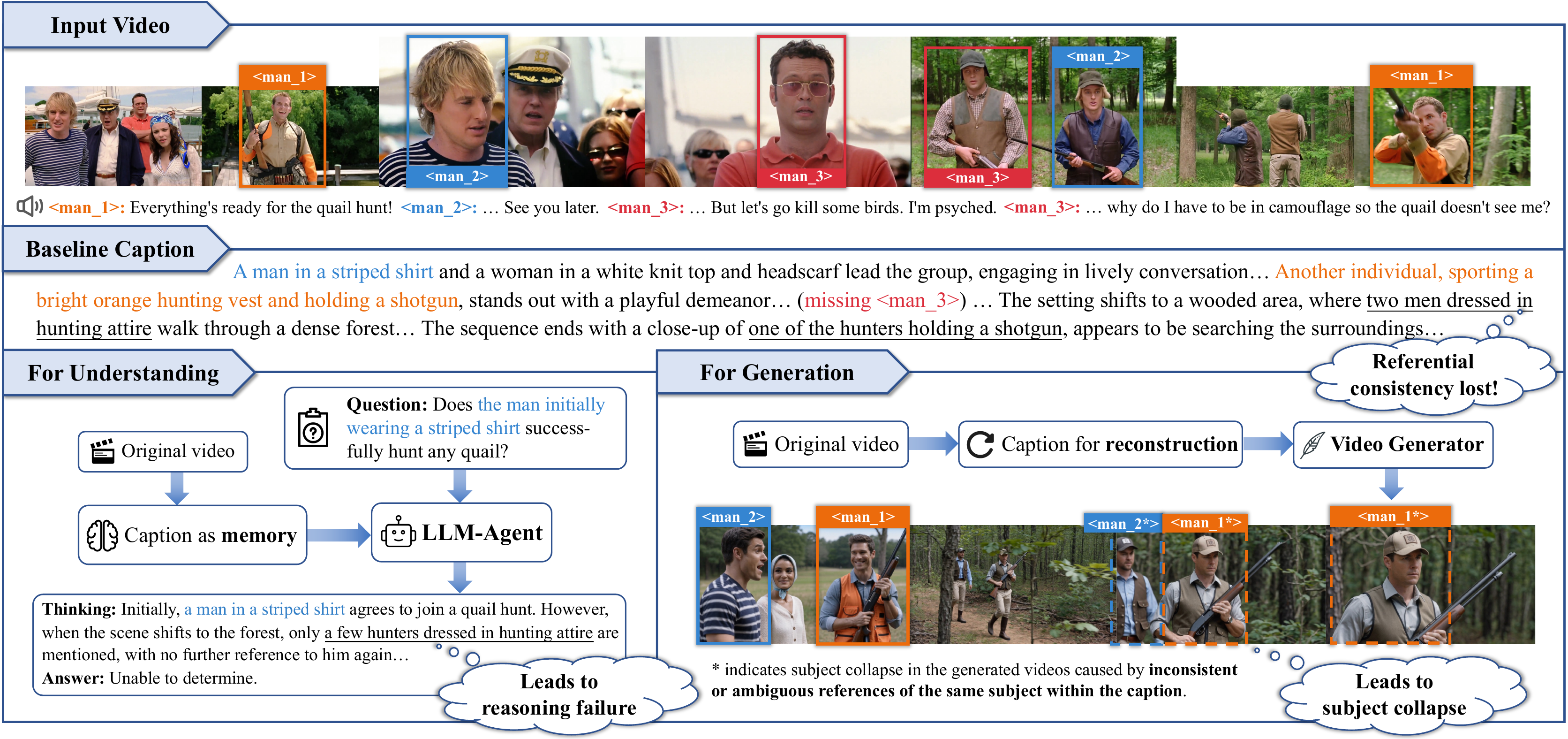}
    \caption{The impact of ambiguous or inconsistent subject references. In the latter half of the baseline caption, the model fails to maintain consistent references to the previously mentioned subject (underlined). Such referential inconsistency degrades downstream performance, leading to reasoning failure when the caption serves as memory for LLM agents in understanding tasks, and causing subject collapse during video reconstruction in generation tasks.}
    \label{fig:teaser}
\end{figure*}

Despite the broad utility of video captioning, current mainstream evaluation benchmarks~\citep{wang2024tarsier, chai2024auroracap, tang2025video} generally suffer from limitations such as (1) restricted video durations; (2) homogeneous content genres; and (3) a lack of scene transitions. The first two constraints prevent existing benchmarks from comprehensively and objectively assessing models' captioning capabilities across varying temporal scales and dynamic real-world environments. Furthermore, the absence of scene changes obscures a critical challenge: maintaining consistent subject references throughout the generated captions, a difficulty significantly amplified by dynamic scene transitions. As illustrated in Figure~\ref{fig:teaser}, ambiguous or inconsistent references to the same subject can severely mislead downstream understanding and generation tasks, such as causing reasoning failure when used as the memory of an LLM agent, or leading to subject collapse during video reconstruction. Consequently, strong performance on existing benchmarks often fails to translate into robust real-world capabilities, hindering researchers from accurately identifying true performance boundaries and conducting targeted optimizations.

To better reflect real-world model performance, an ideal video captioning benchmark should satisfy several essential criteria. At the data level, it should include videos featuring extended temporal spans, diverse domains, and dynamic scene transitions that mirror realistic visual complexity. At the evaluation level, beyond overall caption quality, it should also focus on subject referential consistency, which is particularly challenged by these scene transitions. Additionally, given that video captioning models are commonly developed under either audiovisual or visual-only assumptions, a more comprehensive benchmark should be modality-flexible to support evaluations across both settings.

Motivated by these considerations, we introduce CapRiCorn-1K, the first benchmark dedicated to evaluating video \underline{Cap}tioning and subject \underline{R}eferent\underline{i}al \underline{Con}sistency across long temporal horizons and diverse video scenarios. As detailed in Table~\ref{tab:benchmark_comparison}, CapRiCorn-1K comprises 1,000 manually collected videos featuring dynamic scene transitions. In addition to evaluating overall caption quality, we further introduce a novel metric to quantitatively measure subject referential consistency within generated captions. Furthermore, CapRiCorn-1K supports unified evaluation under both audiovisual (default) and visual-only (CapRiCorn-1K-V) settings.

Extensive experiments on CapRiCorn-1K reveal that current models fall short of generating accurate and comprehensive captions while maintaining consistent subject references. Notably, the performance of open-source models degrades significantly as video duration scales up. To validate the reliability of our benchmark, we employ these captions both as memory for LLM-based agents and as intermediate representations for video reconstruction. Experimental results demonstrate that caption quality evaluated on CapRiCorn-1K correlates strongly with performance in downstream understanding and generation tasks.

Our contributions are summarized as follows:
\begin{itemize}[leftmargin=*]
    \item We introduce CapRiCorn-1K, the first benchmark designed to evaluate video captioning and subject referential consistency across extended temporal horizons, diverse domains, and dynamic scene transitions, enabling a more faithful and comprehensive assessment of captioning performance under both audiovisual and visual-only settings.
    
    \item Through extensive experiments, we demonstrate that existing models generally struggle to generate accurate and comprehensive captions while maintaining consistent subject references. As video duration increases, both overall caption quality and subject referential consistency exhibit a noticeable decline among open-source models.
    
    \item By leveraging captions as memory for LLM-based agents and as intermediate representations for video reconstruction, we show that caption quality, as evaluated on CapRiCorn-1K, strongly correlates with downstream performance in both understanding and generation tasks.
\end{itemize}

\begin{table*}[t]
    \centering
    \resizebox{\linewidth}{!}{
    \begin{tabular}{l|c cccc cccc}
    \toprule
    \multirow{2}{*}[-0.1cm]{\textbf{Benchmark}} & \multirow{2}{*}[-0.1cm]{\textbf{Modality}} & \multirow{2}{*}[-0.1cm]{\textbf{\# Videos}} & \multicolumn{3}{c}{\textbf{Video Duration}} & \multirow{2}{*}[-0.1cm]{\textbf{\makecell{Diverse\\Sources}}} & \multirow{2}{*}[-0.1cm]{\textbf{\makecell{Newly\\Collected}}} & \multirow{2}{*}[-0.1cm]{\textbf{\makecell{Scene\\Trans.}}} & \multirow{2}{*}[-0.1cm]{\textbf{\makecell{Sbj. Ref.\\Consist.}}} \\
    \cmidrule(lr){4-6}
    ~ & ~ & ~ & \textbf{Min.} & \textbf{Avg.} & \textbf{Max.} & ~ & ~ & ~ \\
    \midrule
    DREAM-1K~\citep{wang2024tarsier} & V & 1,000 & 1~s & 9~s & 49~s & \icoyes & \icono & \icono & \icono \\ 
    VDC~\citep{chai2024auroracap} & V & 1,027 & 8~s & 28~s & 163~s & \icoyes & \icono & \icono & \icono \\
    CaReBench~\citep{xu2024carebench} & V & 1,000 & 1~s & 14~s & 124~s & \icoyes & \icono & \icono & \icono \\
    VidCapBench~\citep{chen2025vidcapbench} & V & 643 & 4~s & 10~s & 14~s & \icoyes & \textcolor{BrickRed}{Partial} & \icono & \icono \\
    \midrule
    SALMONN-2 testset~\citep{tang2025video} & A + V & 483 & 31~s & 51~s & 60~s & \icono & \textcolor{gray}{Unknown} & \textcolor{BrickRed}{Partial} & \icono \\
    UGC-VideoCap~\citep{wu2025ugc} & A + V & 1,000 & 8~s & 24~s & 60~s & \icono & \icoyes & \icono & \icono \\
    Omni-Cloze~\citep{ma2025omni} & A + V & 2,320 & 0~s & 34~s & 60~s & \icoyes & \icono & \icono & \icono \\
    \midrule
    CapRiCorn-1K (Ours) & V / (A+V) & 1,000 & 15~s & 252~s & 600~s & \icoyes & \icoyes & \icoyes & \icoyes \\
    \bottomrule
    \end{tabular}
    }
    \caption{Comparison with widely-used video captioning benchmarks. Key dimensions include: evaluation modality (\textbf{Modality}, ``A'' for audio and ``V'' for visual); total number of videos (\textbf{\# Videos}); video duration statistics (\textbf{Min., Avg., Max.}); diversity of video sources (\textbf{Diverse Sources}); whether the videos are independently collected rather than sampled from existing public datasets (\textbf{Newly Collected}); the presence of scene transitions in most videos (\textbf{Scene Trans.}); and the assessment of subject referential consistency in captions (\textbf{Sbj. Ref. Consist.}).}
    \label{tab:benchmark_comparison}
\end{table*}

\section{Related Work}
\subsection{Audiovisual Video Captioning}
The rapid advancement of audiovisual understanding models~\citep{cheng2024videollama, hou2024toward, panagopoulou2023x, shu2025audio, sun2024video, ye2024cat} has catalyzed remarkable progress in audiovisual video captioning. Recent efforts have explored various complementary directions: video-SALMONN-2~\citep{tang2025video}, UGC-VideoCaptioner~\citep{wu2025ugc}, and Omni-Captioner~\citep{ma2025omni} prioritize audiovisual information comprehensiveness; AVoCaDO~\citep{chen2025avocado} focuses on temporal coherence across audiovisual streams; DiaDem~\citep{chen2026diadem} and D-ORCA~\citep{tang2026d} emphasize the fidelity of dialogue descriptions; StoryTeller~\citep{he2024storyteller} incorporates movie cast lists as auxiliary inputs to link dialogue with characters; and several recent studies~\citep{li2026towards, yao2026timechat, geng2025longvale, team2026script, pu2026omniscript} explore structured, time-aware captioning.

Despite model-level advancements, current evaluation benchmarks lag behind, failing to adequately capture real-world complexity. As detailed in Table~\ref{tab:benchmark_comparison}, most existing benchmarks are restricted by limited video durations, narrow domain diversity, and a lack of scene transitions. Such limitations hinder reliable evaluation in dynamic real-world scenarios, thereby impeding the iterations of captioning models toward practical deployment. To bridge this gap, we introduce CapRiCorn-1K, a comprehensive benchmark designed to evaluate video captioning over extended temporal horizons, diverse video domains, and rich scene transitions.

\subsection{Visual-Only Video Captioning}
In the visual-only domain, most existing works~\citep{hu2024fiova, xue2025progress, chen2025vidbridge} have primarily focused on short-video captioning. OwlCap~\citep{zhong2026owlcap} and the Tarsier series~\citep{wang2024tarsier, yuan2025tarsier2} construct large-scale, high-quality datasets to enable the generation of detailed captions that effectively balance dynamic motion and static visual details. AuroraCap~\citep{chai2024auroracap} reduces the input sequence length through token merging while maintaining caption quality.

Regarding long-video captioning, existing works primarily adopt a bottom-up paradigm~\citep{islam2024video, wei2025longcaptioning, chu2025fine}, where videos are first segmented into shorter clips for localized captioning before global aggregation. On the evaluation side, LongCaption-Bench~\citep{wei2025longcaptioning} pioneers the assessment of detailed long-video captioning by measuring caption length, overall quality, and video-caption relevance. Subsequently, RICE-Benchmark~\citep{yang2025addressing} explores the evaluation of identity-matching. However, it only annotates 30 frame indices for subjects in a long video, and such coarse-grained annotations may lead to artificially inflated recall and underestimated precision. In addition, both benchmarks rely on direct LLM-based scoring for caption quality evaluation, which offers limited interpretability. Furthermore, neither benchmark has been open-sourced, restricting their utility in guiding the iterations of long-video captioning models. In contrast, CapRiCorn-1K provides a fine-grained evaluation framework that jointly measures caption quality and subject referential consistency based on video keypoints. Crucially, CapRiCorn-1K will be fully open-sourced to facilitate future research.

\begin{figure*}[t]
    \centering
    \includegraphics[width=\linewidth]{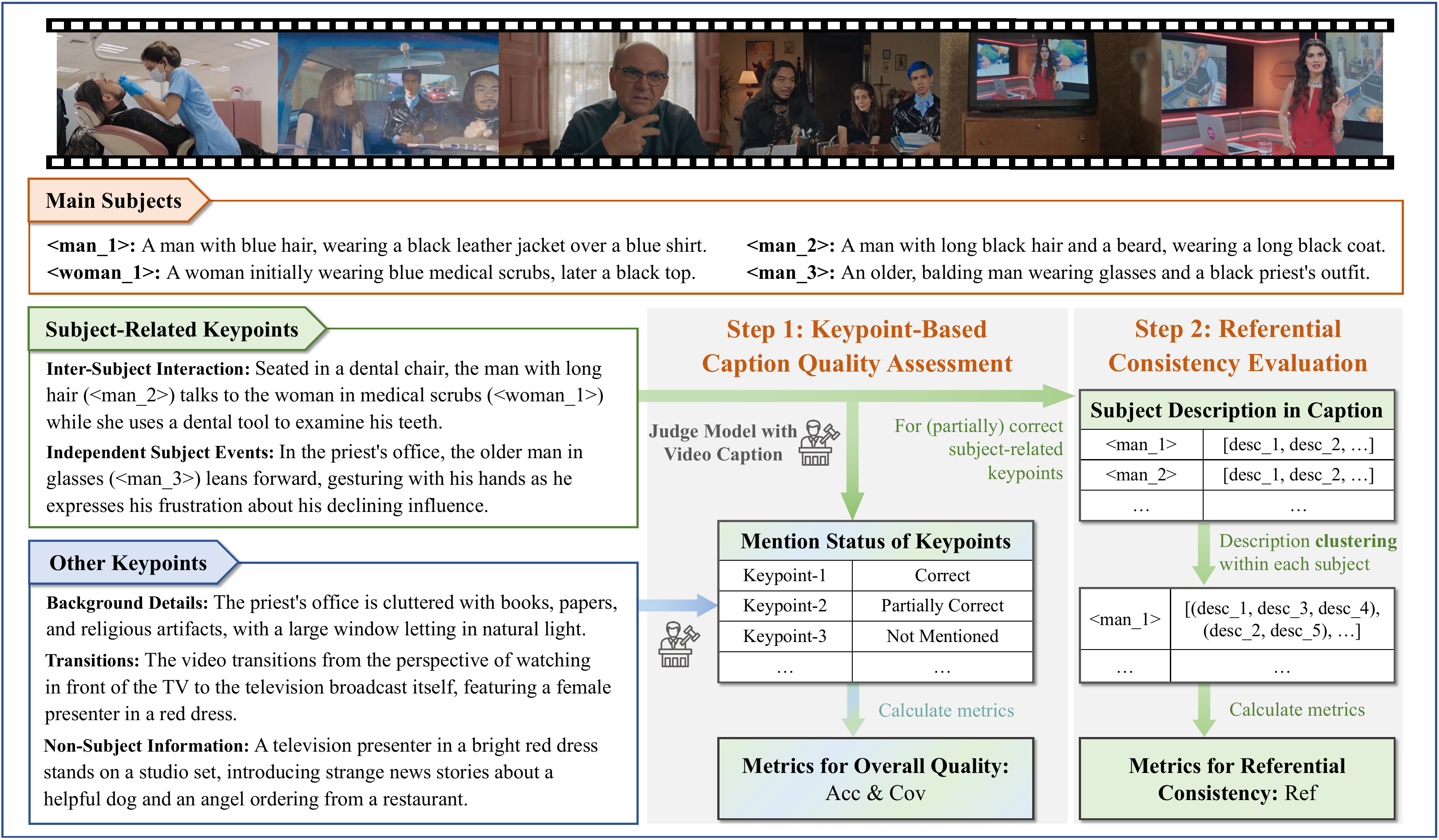}
    \caption{Evaluation pipeline of CapRiCorn-1K: (1) determining the mention status of all keypoints to assess overall caption quality (Acc \& Cov); and (2) extracting the localized subject descriptions from the caption for all mentioned subject-related keypoints, which are then clustered to assess referential consistency (Ref).}
    \label{fig:eval}
\end{figure*}
\section{CapRiCorn-1K}
\subsection{Overview}
As a benchmark tailored for evaluating video captioning over extended temporal spans, CapRiCorn-1K aims to comprehensively assess both the overall caption quality and the referential consistency of recurring subjects across diverse video scenarios. In this section, we detail the evaluation protocols, video collection criteria, annotation methodology, and statistical characteristics of CapRiCorn-1K.

\subsection{Evaluation Protocols}
Inspired by the video-SALMONN-2 testset~\citep{tang2025video}, we first decompose each video into a sequence of categorized keypoints. A judge model (GPT-4.1) is then employed to verify the mention status of each keypoint within the generated caption, thereby evaluating the overall captioning quality. Furthermore, to assess the model's ability to maintain referential consistency for the same subject over long contexts, we utilize these keypoints as anchors to extract corresponding subject descriptions from the caption. The judge model then determines whether the descriptions associated with the same ground-truth subject remain referentially consistent within the caption context, thereby deriving a subject referential consistency score. The complete evaluation pipeline is illustrated in Figure~\ref{fig:eval}, with formal definitions provided below.

\subsubsection{Overall Captioning Quality}
For a given video, we first manually identify a set of ground-truth subjects $\mathcal{S}=\{s_1, s_2, \dots, s_m\}$ and partition the video into a set of keypoints $\mathcal{K}=\{k_1, k_2, \dots, k_n\}$. As detailed in Section~\ref{sec:anno}, these keypoints are classified into five categories: inter-subject interaction ($\mathcal{K}_{\text{inter}}$), independent subject events ($\mathcal{K}_{\text{indep}}$), background details ($\mathcal{K}_{\text{bg}}$), transitions ($\mathcal{K}_{\text{trans}}$), and non-subject information ($\mathcal{K}_{\text{non}}$).

The judge model evaluates the overall captioning quality by assigning a discrete mention status $y_i \in \{\text{correct}, \text{partial}, \text{none}\}$ to each keypoint $k_i$, corresponding to ``correctly mentioned'', ``partially mentioned or containing errors'', and ``not mentioned''. Let $\mathcal{K}^{\text{correct}} = \{k_i \in \mathcal{K} \mid y_i = \text{correct}\}$ and $\mathcal{K}^{\text{partial}} = \{k_i \in \mathcal{K} \mid y_i = \text{partial}\}$, we define Accuracy ($\text{Acc}$) and Coverage ($\text{Cov}$) to measure the overall caption quality as follows:
\begin{equation}
\text{Acc} = \frac{|\mathcal{K}^{\text{correct}}|}{|\mathcal{K}|}, \quad
\text{Cov} = \frac{|\mathcal{K}^{\text{correct}}| + |\mathcal{K}^{\text{partial}}|}{|\mathcal{K}|}.
\end{equation}

\subsubsection{Subject Referential Consistency}
To assess the referential consistency for a specific subject $s_j$, we utilize keypoints as anchors to extract subject descriptions from the caption, and subsequently determine whether these descriptions co-refer to the same subject contextually. Formally, let $\mathcal{K}_{s_j} \subseteq \mathcal{K}_{\text{inter}} \cup \mathcal{K}_{\text{indep}}$ denote the set of subject-related keypoints associated with $s_j$. For each keypoint $k_i \in \mathcal{K}_{s_j}$ that has been judged as correctly or partially mentioned (i.e., with mention status $y_i \in \{\text{correct}, \text{partial}\}$), the judge model extracts the corresponding localized subject description from the caption. This yields a set of caption-derived subject descriptions belonging to $s_j$, denoted as $\mathcal{D}_{s_j} = \{d_{j,1}, d_{j,2}, \dots, d_{j,N_j}\}$.

Notably, a subject's appearance (e.g., clothing) may vary across different scenes. Therefore, evaluating referential consistency relying solely on the isolated semantics of the descriptions in $\mathcal{D}_{s_j}$ is inadequate. Considering that continuous subject tracking requires the caption to explicitly document these appearance variations, the judge model is instructed to perform co-reference clustering on $\mathcal{D}_{s_j}$ based on the caption context, resulting in disjoint co-reference partitions $\mathcal{P}_{s_j} = \{P_{j,1}, P_{j,2}, \dots, P_{j,C_j}\}$.

A naive approach to quantifying subject referential consistency is to rely on the number of clusters, $|\mathcal{P}_{s_j}|$, where more clusters indicate lower consistency. However, this could introduce bias by ignoring the size distribution among clusters. For instance, given $|\mathcal{D}_{s_j}| = 6$, a cluster size distribution of $\{1, 1, 4\}$ inherently reflects higher consistency than $\{1, 2, 3\}$, despite both yielding $|\mathcal{P}_{s_j}| = 3$.

To mitigate this bias, we draw inspiration from the Rand Index~\citep{rand1971objective} and define the subject-level referential consistency score ($\text{Ref}_j$) as the ratio between: (i) the number of pairwise combinations of subject descriptions in $\mathcal{D}_{s_j}$ that belong to the same cluster, and (ii) the total number of pairwise combinations among all keypoints in $\mathcal{K}_{s_j}$. Crucially, by utilizing $|\mathcal{K}_{s_j}|$ rather than $|\mathcal{D}_{s_j}|$ in the denominator, the metric explicitly penalizes models that inflate consistency scores by generating overly concise captions (i.e., where $|\mathcal{D}_{s_j}| \ll |\mathcal{K}_{s_j}|$). For subjects with $|\mathcal{K}_{s_j}| \ge 2$ (a condition met by all subjects in CapRiCorn-1K), $\text{Ref}_j$ is formulated as:
\begin{equation} \text{Ref}_j = \frac{\sum_{c=1}^{|\mathcal{P}_{s_j}|} \binom{|P_{j,c}|}{2}}{\binom{|\mathcal{K}_{s_j}|}{2}},
\end{equation}
where $|P_{j,c}|$ denotes the number of descriptions within the $c$-th cluster. Finally, the video-level referential consistency score ($\text{Ref}$) is computed by averaging across all subjects:
\begin{equation}
\text{Ref} = \frac{1}{|\mathcal{S}|} \sum_{s_j \in \mathcal{S}} \text{Ref}_j.
\end{equation}

\subsection{Video Collection}
\label{sec:video_collection}
Unlike many existing benchmarks that sample evaluation subsets from established datasets, we manually collect and process videos from the Internet.

To enable a more comprehensive assessment of video captioning performance across diverse scenarios, CapRiCorn-1K is carefully curated to cover extended and balanced temporal spans, as well as a wide variety of video content. Regarding video duration, we substantially broaden the temporal scope compared with mainstream benchmarks, selecting videos ranging from 15 seconds to 10 minutes. Videos shorter than 15 seconds are excluded because they typically contain limited dynamics. In terms of content diversity, we collect videos from eight major categories to ensure broad domain coverage: Relationship, Youth, Entertainment, History, Family, Lifestyle, Fantasy, and Mystery. Each major category is further divided into multiple fine-grained subcategories, as detailed in Table~\ref{tab:full_category}.

Furthermore, to better reflect real-world dynamics and to more rigorously evaluate referential consistency for the same subject over time, each video is required to contain at least one scene transition, rather than merely camera-shot changes within a single scene. This criterion forces models to rely on genuine, identity-related visual cues rather than relative spatial positioning to track subjects. To introduce an additional layer of complexity, approximately 40\% of the collected videos feature subjects undergoing clothing changes.

Finally, we impose additional requirements such as video resolution to guarantee video quality. More details are provided in Appendix~\ref{app:collect_detail}.

\subsection{Data Annotation}
\label{sec:anno}
Following video collection, we conduct rigorous manual annotation. Compared to automated annotation, whose scope and accuracy are inherently limited by the capabilities of the underlying model, manual annotation better reflects real-world requirements and yields more reliable ground truth.

To support the evaluation of subject referential consistency in video captioning, we first identify the primary subjects within each video. A subject is defined as a character who actively drives the storyline and significantly contributes to the narrative progression. Two annotators independently identify the subjects and cross-validate their results, with discrepancies resolved by a senior annotator.

Subsequently, we annotate keypoints across five categories to comprehensively evaluate overall caption quality and subject referential consistency:
\begin{itemize}[leftmargin=*]
    \item \textbf{Inter-Subject Interactions:} Interactions among multiple subjects;
    \item \textbf{Independent Subject Events:} Actions or events performed by a single subject;
    \item \textbf{Background Details:} Contextual information such as visual background elements, ambient sounds, and other environmental cues;
    \item \textbf{Transitions:} Scene transitions, camera shifts, and environmental changes;
    \item \textbf{Non-Subject Information:} Salient events or details not directly related to the primary subjects.
\end{itemize}

To balance annotation granularity with evaluation cost, three annotators independently identify approximately 40 salient keypoints per video. Two senior annotators then each review the three annotation sets and select keypoints exhibiting high consensus and critical narrative importance. Finally, a lead expert consolidates and verifies these two refined sets to form the final keypoint collection.

Furthermore, to cater to visual-only captioning models, two additional annotators filter and cross-validate this final collection to derive a vision-only keypoint subset, denoted as CapRiCorn-1K-V. Disagreements during this stage are likewise resolved by a senior annotator. Detailed information regarding the annotators is provided in Appendix~\ref{app:human_anno}.

\subsection{Benchmark Statistics}
\begin{figure}
    \centering
    \includegraphics[width=\linewidth]{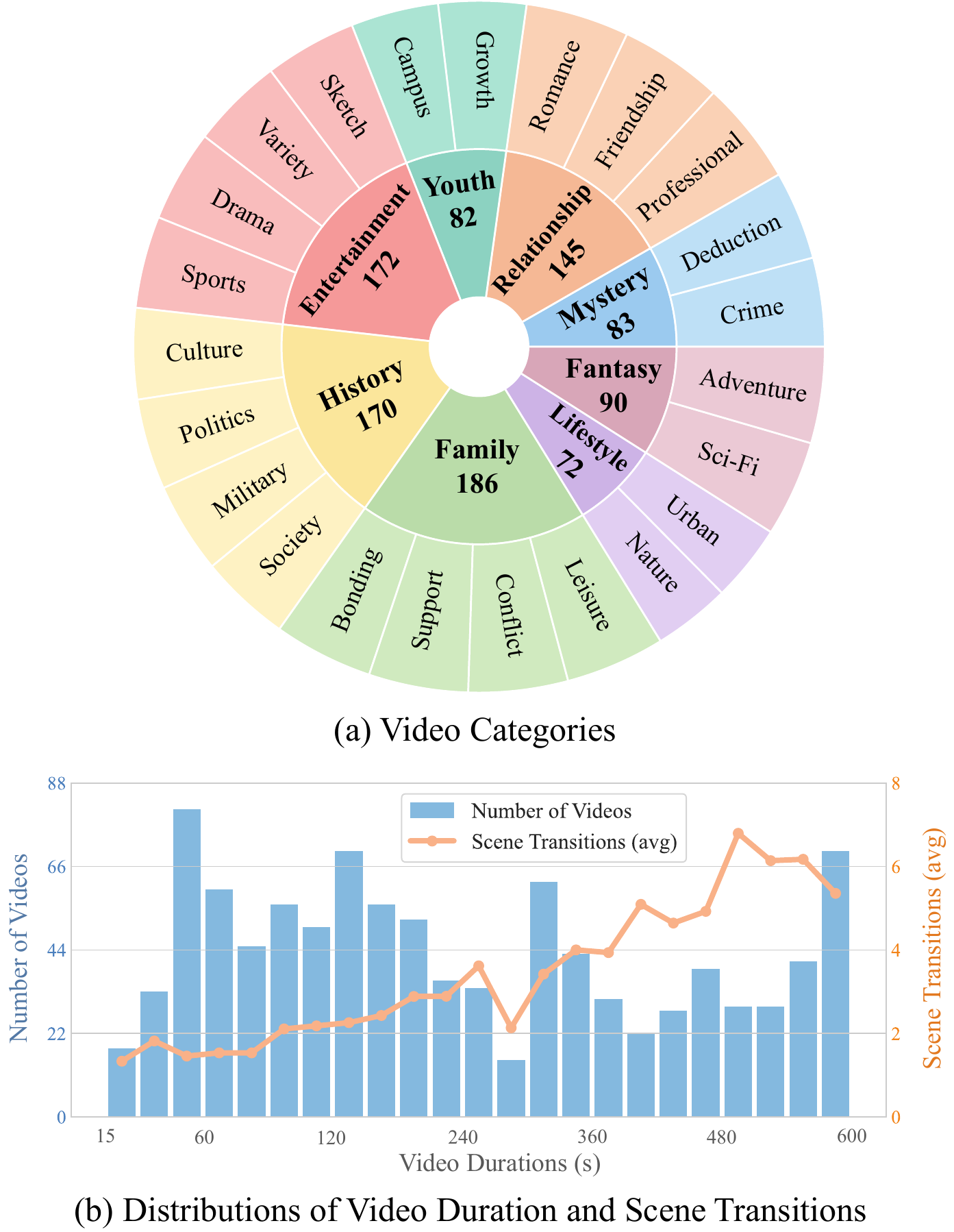}
    \caption{Statistics of CapRiCorn-1K: (a) Diverse category distribution; and (b) Balanced duration distribution with rich scene transitions.}
    \label{fig:statistics}
\end{figure}

As shown in Table~\ref{tab:benchmark_comparison} and Figure~\ref{fig:statistics}, CapRiCorn-1K comprises 1,000 newly collected videos evenly distributed across eight major categories. Video durations range uniformly from 15 to 600 seconds, yielding an average length of 252 seconds. Notably, each video features an average of 3.1 scene transitions, with the transition density scaling with video duration, reflecting the high dynamics of our benchmark. In terms of annotations, each video is meticulously labeled with an average of 4.4 subjects, 21.5 salient subject-related keypoints, and 14.9 salient keypoints of other types.

\section{Experiments}
Our evaluation adheres to the official protocols of each model by default. When such protocols are unavailable, given the substantial length of the videos, we uniformly sample frames up to the maximum context window supported by the model while preserving sufficient frame resolution. More implementation details are provided in Appendix~\ref{app:imple_detail}.

\begin{table*}[t]
    \centering
    \resizebox{\linewidth}{!}{
    \begin{tabular}{l@{\hspace{-4pt}}c|>{\columncolor{gray!10}}c>{\columncolor{gray!10}}c>{\columncolor{gray!10}}c|ccc|ccc|ccc|ccc}
    \toprule
    \multirow{2}{*}{\textbf{Model}} & \multirow{2}{*}{\textbf{Size}} & \multicolumn{3}{>{\columncolor{gray!10}}c|}{\textbf{Overall}} & \multicolumn{3}{c|}{\textbf{(0, 2]~min}} & \multicolumn{3}{c|}{\textbf{(2, 5]~min}} & \multicolumn{3}{c|}{\textbf{(5, 8]~min}} & \multicolumn{3}{c}{\textbf{(8, 10]~min}} \\
    ~ & ~ & \textbf{Acc} & \textbf{Cov} & \textbf{Ref} & \textbf{Acc} & \textbf{Cov} & \textbf{Ref} & \textbf{Acc} & \textbf{Cov} & \textbf{Ref} & \textbf{Acc} & \textbf{Cov} & \textbf{Ref} & \textbf{Acc} & \textbf{Cov} & \textbf{Ref} \\
    \midrule
    Gemini-3.1-Pro & - & \textbf{42.5} & \textbf{53.3} & \underline{39.1} & 40.9 & 53.4 & \underline{42.4} & \textbf{44.7} & \textbf{54.8} & \textbf{40.4} & \textbf{42.4} & \textbf{52.7} & \underline{35.3} & \textbf{42.2} & \textbf{51.6} & \textbf{35.4} \\
    Gemini-3-Flash & - & \underline{41.5} & \underline{52.8} & \textbf{39.6} & \underline{42.8} & \underline{55.9} & \textbf{46.3} & \underline{42.1} & \underline{52.9} & \underline{38.1} & \underline{41.1} & \underline{51.3} & \textbf{36.6} & \underline{38.5} & \underline{48.3} & \underline{32.3} \\
    \midrule
    Qwen2.5-Omni & 3B & 4.1 & 11.6 & 0.5 & 5.8 & 15.9 & 1.2 & 4.1 & 11.5 & 0.2 & 2.6 & 8.4 & 0.1 & 2.3 & 7.1 & 0.1 \\
    video-SALMONN-2+ & 3B & 9.4 & 19.0 & 1.1 & 11.8 & 23.5 & 1.9 & 9.1 & 18.1 & 0.8 & 8.1 & 16.8 & 0.6 & 6.5 & 14.1 & 0.4 \\
    UGC-VideoCaptioner & 3B & 11.8 & 21.8 & 3.6 & 17.4 & 30.6 & 7.4 & 11.0 & 20.1 & 2.1 & 8.3 & 17.0 & 1.4 & 6.5 & 13.2 & 0.9 \\
    ASID-Captioner & 3B & 12.8 & 23.2 & 7.0 & 21.7 & 37.2 & 14.9 & 11.9 & 21.3 & 5.2 & 6.3 & 13.9 & 1.7 & 4.5 & 9.8 & 0.7 \\
    \midrule
    ARC-Qwen-Video-Narrator & 7B & 2.3 & 3.2 & 0.6 & 4.6 & 6.7 & 1.5 & 1.5 & 0.2 & 0.2 & 1.0 & 1.2 & 0.1 & 0.6 & 0.8 & 0.0 \\
    Qwen2.5-Omni & 7B & 5.1 & 13.2 & 0.6 & 6.7 & 17.4 & 1.2 & 5.9 & 13.7 & 0.5 & 3.4 & 10.0 & 0.3 & 2.6 & 8.5 & 0.1 \\
    OmniVinci & 9B & 5.9 & 13.3 & 1.2 & 9.9 & 21.2 & 2.5 & 5.3 & 12.0 & 0.6 & 3.2 & 8.3 & 0.4 & 2.5 & 6.2 & 0.5 \\
    ARC-Qwen-Video & 7B & 6.9 & 10.9 & 2.0 & 9.2 & 15.3 & 3.4 & 7.2 & 11.0 & 1.8 & 5.8 & 9.0 & 1.4 & 2.8 & 4.4 & 0.4 \\
    video-SALMONN-2+ & 7B & 9.3 & 18.7 & 1.4 & 12.1 & 24.0 & 2.3 & 9.1 & 17.5 & 1.0 & 7.2 & 15.6 & 0.6 & 6.7 & 13.8 & 1.0 \\
    ASID-Captioner & 7B & 18.9 & 31.1 & 12.9 & 30.2 & 47.3 & 26.3 & 18.6 & 30.4 & 10.3 & 10.5 & 19.3 & 3.8 & 7.7 & 14.9 & 1.9 \\
    video-SALMONN-2 & 7B & 22.5 & 37.6 & 11.3 & 27.6 & 46.0 & 18.2 & 23.2 & 37.6 & 10.6 & 19.9 & 33.2 & 7.1 & 14.3 & 26.2 & 4.0 \\
    DiaDem & 7B & 24.6 & 35.8 & 14.5 & 40.0 & 54.7 & 31.3 & 23.1 & 33.2 & 10.2 & 13.6 & 23.0 & 3.3 & 10.3 & 18.3 & 2.0 \\
    AVoCaDO & 7B & 28.8 & 41.9 & 18.4 & \textbf{43.7} & \textbf{60.6} & 36.6 & 29.8 & 42.4 & 15.8 & 17.0 & 27.5 & 5.2 & 12.9 & 22.3 & 3.1 \\
    \midrule
    Qwen3-Omni-Instruct & 30B-A3B & 10.3 & 20.2 & 1.6 & 13.3 & 24.6 & 2.5 & 10.6 & 20.3 & 1.6 & 8.0 & 17.6 & 0.9 & 6.5 & 14.5 & 0.7 \\
    Qwen3-Omni-Captioner & 30B-A3B & 14.3 & 27.5 & 4.1 & 18.1 & 33.5 & 7.0 & 14.4 & 27.4 & 3.4 & 11.4 & 23.0 & 2.3 & 10.2 & 21.4 & 1.9 \\
    video-SALMONN-2+ & 72B & 11.5 & 21.5 & 1.9 & 14.6 & 26.8 & 3.0 & 10.9 & 20.1 & 1.4 & 9.6 & 18.8 & 1.3 & 8.6 & 16.5 & 1.1 \\
    \bottomrule
    \end{tabular}
    }
    \caption{Evaluation results of audiovisual captioning models on CapRiCorn-1K.}
    \label{tab:result_av}
\end{table*}

\begin{table*}[t]
    \centering
    \resizebox{\linewidth}{!}{
    \begin{tabular}{l@{\hspace{-4pt}}c|>{\columncolor{gray!10}}c>{\columncolor{gray!10}}c>{\columncolor{gray!10}}c|ccc|ccc|ccc|ccc}
    \toprule
    \multirow{2}{*}{\textbf{Model}} & \multirow{2}{*}{\textbf{Size}} & \multicolumn{3}{>{\columncolor{gray!10}}c|}{\textbf{Overall}} & \multicolumn{3}{c|}{\textbf{(0, 2]~min}} & \multicolumn{3}{c|}{\textbf{(2, 5]~min}} & \multicolumn{3}{c|}{\textbf{(5, 8]~min}} & \multicolumn{3}{c}{\textbf{(8, 10]~min}} \\
    ~ & ~ & \textbf{Acc} & \textbf{Cov} & \textbf{Ref} & \textbf{Acc} & \textbf{Cov} & \textbf{Ref} & \textbf{Acc} & \textbf{Cov} & \textbf{Ref} & \textbf{Acc} & \textbf{Cov} & \textbf{Ref} & \textbf{Acc} & \textbf{Cov} & \textbf{Ref} \\
    \midrule
    Tarsier2 & 7B & 7.5 & 18.8 & 4.6 & 9.2 & 23.6 & 7.0 & 7.5 & 18.7 & \textbf{4.3} & 6.8 & 15.9 & \textbf{3.6} & 5.1 & 12.9 & \textbf{1.8} \\
    MiMo-VL & 7B & 11.7 & 23.7 & 1.5 & 15.6 & 30.2 & 2.8 & 11.9 & 24.4 & 1.4 & 9.6 & 19.9 & 0.5 & 6.2 & 14.6 & 0.3 \\
    Qwen3.5 & 9B & 10.7 & 24.7 & 3.1 & 15.2 & 31.9 & 6.3 & 10.5 & 25.0 & 2.5 & 7.3 & 19.2 & 1.4 & 6.2 & 16.5 & 0.2 \\
    InternVL3.5 & 8B & 13.2 & \underline{28.2} & \textbf{5.4} & 18.4 & 35.3 & \underline{9.8} & \underline{12.4} & \underline{27.8} & \underline{4.1} & \underline{10.1} & \underline{23.3} & \underline{3.0} & \underline{8.3} & \textbf{20.7} & \underline{1.7} \\
    Qwen3-VL & 8B & \textbf{15.8} & \textbf{30.2} & \underline{5.1} & \textbf{23.3} & \textbf{40.6} & \textbf{10.8} & \textbf{14.3} & \textbf{28.7} & 3.0 & \textbf{11.2} & \textbf{23.7} & 1.9 & \textbf{9.0} & \underline{20.1} & 1.1 \\
    \midrule
    Qwen3.6 & 27B & \underline{13.4} & 27.8 & 3.1 & \underline{19.0} & \underline{36.5} & 6.7 & 12.3 & 26.9 & 1.8 & 9.9 & 22.1 & 1.1 & 8.1 & 19.0 & 0.8 \\
    Qwen3.6 & 35B-A3B & 11.7 & 25.8 & 2.9 & 16.2 & 32.6 & 5.2 & 11.1 & 25.5 & 2.6 & 8.6 & 21.5 & 1.1 & 7.8 & 18.3 & 1.1 \\
    Qwen3.5 & 122B-A10B & 11.7 & 25.6 & 2.6 & 15.2 & 31.9 & 5.2 & 12.1 & 25.6 & 1.7 & 9.0 & 21.1 & 1.2 & 7.7 & 18.6 & 0.5 \\
    \bottomrule
    \end{tabular}
    }
    \caption{Evaluation results of visual-only captioning models on CapRiCorn-1K-V.}
    \label{tab:result_v}
\end{table*}

\subsection{Captioning Models}
For the default audiovisual setting (CapRiCorn-1K), we assess the Gemini series~\citep{comanici2025gemini}, Qwen-Omni series~\citep{xu2025qwen2, xu2025qwen3}, video-SALMONN-2 series~\citep{tang2025video}, ARC-Qwen-Video~\citep{ge2025arc}, OmniVinci~\citep{ye2025omnivinci}, UGC-VideoCaptioner~\citep{wu2025ugc}, AVoCaDO~\citep{chen2025avocado}, DiaDem~\citep{chen2026diadem}, and ASID-Captioner~\citep{li2026towards}.

For the visual-only setting (CapRiCorn-1K-V), we evaluate Tarsier2~\citep{yuan2025tarsier2}, MiMo-VL~\citep{coreteam2025mimovltechnicalreport}, Qwen3-VL~\citep{bai2025qwen3}, InternVL3.5~\citep{wang2025internvl3}, Qwen3.5~\citep{qwen35blog}, and Qwen3.6~\citep{qwen3.6-27b, qwen3.6-35b-a3b}.

\subsection{Main Results}
Tables~\ref{tab:result_av} and~\ref{tab:result_v} present the performance of various audiovisual captioning models on CapRiCorn-1K and vision-only captioning models on CapRiCorn-1K-V, respectively. Our key findings are as follows:

\begin{itemize}[leftmargin=*]
\item \textbf{Performance Gap and Long-Video Robustness.} Existing models generally struggle to generate accurate and comprehensive captions with consistent subject references. Overall, the closed-source Gemini series consistently outperforms open-source models by a large margin, and its captioning performance only degrades marginally as video duration increases. In contrast, open-source models exhibit severe performance drops on longer videos, particularly in maintaining referential consistency.

\item \textbf{Limitations of Existing Benchmarks.} While certain specialized open-source models (e.g., AVoCaDO and DiaDem) achieve overall captioning quality comparable to the Gemini series on short videos (0 to 2 minutes), they lag substantially behind in terms of subject referential consistency and long-video robustness. One possible reason is that these models are primarily optimized for existing benchmarks, which mainly emphasize overall caption quality on short videos, thereby overlooking long-duration videos and subject referential consistency, both of which are more critical in real-world applications.

\item \textbf{Captioning Performance Depends on Multiple Factors.} Although increasing parameter scale yields performance gains within specific model families (e.g., Qwen2.5-Omni, Qwen3.5, and video-SALMONN-2+), larger model size alone does not guarantee superior performance. For instance, despite having only 7B parameters, AVoCaDO substantially outperforms the 72B version of video-SALMONN-2+. This highlights that captioning capability is also influenced by other critical components, such as architectural design, training data distribution and optimization strategies, rather than parameter scale alone.
\end{itemize}

\subsection{Ablation on the Judge Model}
In the main experiments, we adopt GPT-4.1 as the judge model. To account for scenarios where closed-source APIs are unavailable, and to further assess the generalizability of our evaluation protocol across different judge models, we conduct an ablation study by replacing GPT-4.1 with the open-source Qwen3-235B-A22B-Instruct~\citep{yang2025qwen3}. The results are reported in Table~\ref{tab:abla_judge}.

The experimental results reveal that, although the absolute scores produced by different judge models exhibit fluctuations, which may stem from inherent model-specific biases (e.g., Qwen3-235B-A22B-Instruct tending to be more conservative on Accuracy), the relative rankings among the evaluated models remain largely consistent. Specifically, the Pearson correlation coefficients~\citep{benesty2009pearson} between the scores produced by different judge models across the three evaluation metrics reach 0.999, 0.998, and 0.998, respectively ($p < 0.001$), indicating that our evaluation protocol is not strictly dependent on a specific judge model. Instead, as long as the judge model possesses strong capabilities and can deliver stable, fair judgments, it is suitable for integration into CapRiCorn-1K. 

\begin{table}[t]
    \centering
    \resizebox{\linewidth}{!}{
    \begin{tabular}{l|ccc|c@{\hspace{16pt}}cc}
    \toprule
    \multirow{2}{*}{\textbf{Model}} & \multicolumn{3}{c|}{\textbf{GPT-4.1}} & \multicolumn{3}{c}{\textbf{Qwen3-235B-A22B}} \\
    ~ & \textbf{Acc} & \textbf{Cov} & \textbf{Ref} & \textbf{Acc} & \textbf{Cov} & \textbf{Ref} \\
    \midrule
    Gemini-3.1-Pro & 42.5 & 53.3 & 39.1 & 27.2 & 51.6 & 35.4 \\
    Qwen2.5-Omni & 5.1 & 13.2 & 0.6 & 5.3 & 16.8 & 1.3 \\
    Qwen3-Omni-Captioner & 14.3 & 27.5 & 4.1 & 10.7 & 29.1 & 4.7 \\
    ASID-Captioner-7B & 18.9 & 31.1 & 12.9 & 12.7 & 30.4 & 13.0 \\
    AVoCaDO & 28.8 & 41.9 & 18.4 & 18.9 & 41.7 & 18.9 \\
    \bottomrule
    \end{tabular}
    }
    \caption{Ablation on the judge model.}
    \label{tab:abla_judge}
\end{table}

\subsection{Correlation with Downstream Tasks}
\begin{figure}[t]
    \centering
    \includegraphics[width=\linewidth]{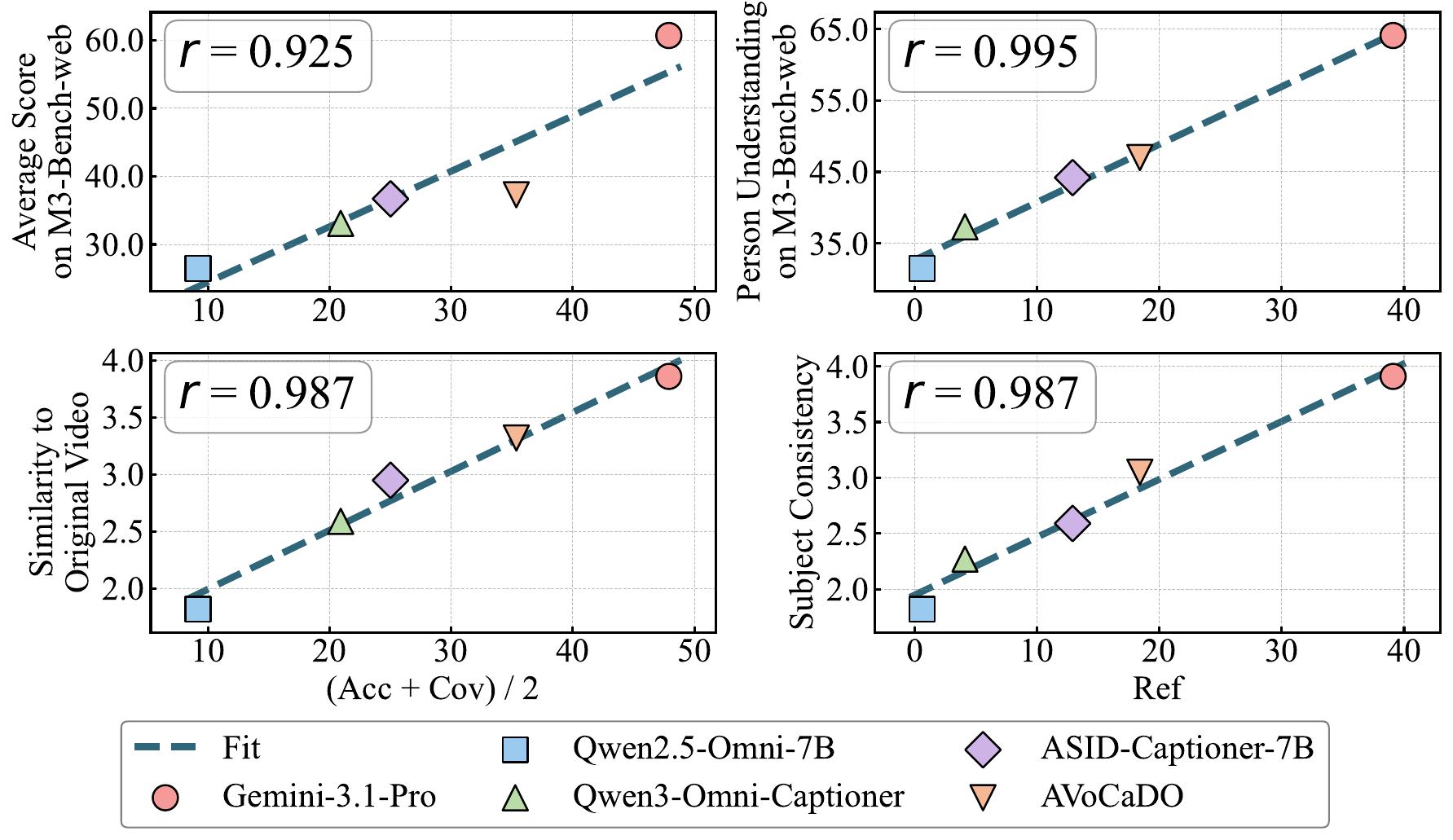}
    \caption{Correlation between evaluation metrics on CapRiCorn-1K with downstream task performance.}
    \label{fig:corr}
\end{figure}

To validate the reliability of our evaluation metrics, we apply the generated captions to downstream understanding and generation tasks, examining the correlation between downstream task performance and our metric scores in Figure~\ref{fig:corr}.

For the understanding task, we adopt M3-Bench-web~\citep{long2025seeing}, a benchmark featuring long videos designed to evaluate the reasoning capabilities of multimodal agents for long-term memory. Following the ``Socratic Models'' paradigm used in M3-Bench-web, we first generate video captions using different captioning models and then supply these captions as memory to a fixed LLM agent (GPT-4.1). Consequently, the reasoning performance of this LLM agent serves as a direct indicator of caption quality. As illustrated in the upper panel of Figure~\ref{fig:corr}, the overall caption quality (measured by Acc and Cov) exhibits a strong correlation with the average score of the LLM agent on M3-Bench-web (upper left), yielding a Pearson correlation coefficient of 0.925. Moreover, the consistency of subject references within the captions (measured by Ref) shows an even stronger correlation with the ``Person Understanding'' subset of M3-Bench-web (upper right), achieving a Pearson correlation coefficient of 0.995.

For the generation task, we randomly sample 50 videos from CapRiCorn-1K and leverage captions generated by different models to reconstruct the original videos using LTX-2.3-22B-dev~\citep{hacohen2025ltx2}. Human evaluators then rate both the similarity between the generated and original videos, as well as the subject consistency within the generated videos, on a scale from 1 to 5. These scores, averaged across three annotators, serve as a reliable proxy for caption quality. The results in the lower panel of Figure~\ref{fig:corr} demonstrate that the overall caption quality is highly correlated with video similarity, while the referential consistency of subjects in the captions aligns strongly with the subject consistency of generated videos, with both Pearson correlation coefficients reaching 0.987.

\subsection{Further Analysis}
Additional investigations regarding the impacts of caption length, input frame count, and input resolution, along with a detailed error analysis, are provided in Appendix~\ref{app:further_analysis}.

\section{Conclusion}
In this paper, we present CapRiCorn-1K, a comprehensive benchmark designed to evaluate video captioning and subject referential consistency across diverse durations and scenarios. To better capture real-world complexity, we manually collect and annotate 1,000 videos spanning long temporal horizons and various domains. Furthermore, we propose a suite of evaluation metrics to assess overall caption quality and subject referential consistency under both audiovisual and vision-only settings. By integrating the generated captions into downstream understanding and generation tasks, we demonstrate that the evaluation results on CapRiCorn-1K exhibit strong correlations with downstream task performance, thereby validating the reliability and practical utility of our benchmark.

\section*{Limitations}
While CapRiCorn-1K significantly extends the video duration compared to existing video captioning benchmarks, its scope remains restricted to videos under 10 minutes. Given that current models still face considerable challenges within this time span, we hope our benchmark serves as a stepping stone, leaving the evaluation of longer videos to future research. Additionally, due to the substantial domain differences between human and non-human subjects, coupled with the prevalence of human-centric content in practical applications (e.g., human-computer interaction and surveillance), our study prioritizes referential consistency in human subjects as a more critical and immediate challenge to address, leaving the exploration of non-human subjects for future work.

\section*{Ethical Considerations}
The videos in CapRiCorn-1K are collected from publicly available online platforms. To strictly adhere to copyright regulations and respect intellectual property rights, our benchmark will be released under highly restrictive licensing terms, allowing its use exclusively for academic research purposes.

\bibliography{custom}

\clearpage
\newpage
\appendix

\begin{center}
\section*{Appendix}
\end{center}

\section{Video Collection Details}
\label{app:collect_detail}
Beyond the requirements on video duration and content diversity discussed in Section~\ref{sec:video_collection}, we further elaborate on our quality control protocols and provide the complete taxonomy of our benchmark.

First, in terms of resolution and visual fidelity, all videos are required to have a minimum resolution of 720p. Videos exhibiting severe artifacts, such as over-sharpening, noticeable mosaic distortion, or audiovisual misalignment, are excluded. Second, to prevent semantic leakage, we filter out videos containing excessive on-screen subtitles, as such text could inadvertently provide cues that confound the evaluation of a model's audiovisual fusion capability. Third, to minimize domain bias, we restrict the collection to at most one video from each distinct source, such as a specific movie, television series, or content creator. Finally, to reduce the risk of data contamination, we not only exclude samples from existing datasets but also prioritize recently published videos. To respect copyright constraints, our benchmark will be released under highly restrictive licensing terms, permitting its use exclusively for academic research purposes.

Due to space constraints in the main text, Figure~\ref{fig:statistics} only illustrates the eight primary domains and their major subcategories. To provide a comprehensive overview, Table~\ref{tab:full_category} provides the full taxonomy of all 36 fine-grained video subcategories included in our collection.

\section{Human Annotators}
\label{app:human_anno}
During the video collection stage, we recruit ten experienced video collectors through a crowdsourcing platform to gather videos from the Internet that satisfy our predefined selection criteria.

\begin{figure*}[t]
    \centering
    \includegraphics[width=\linewidth]{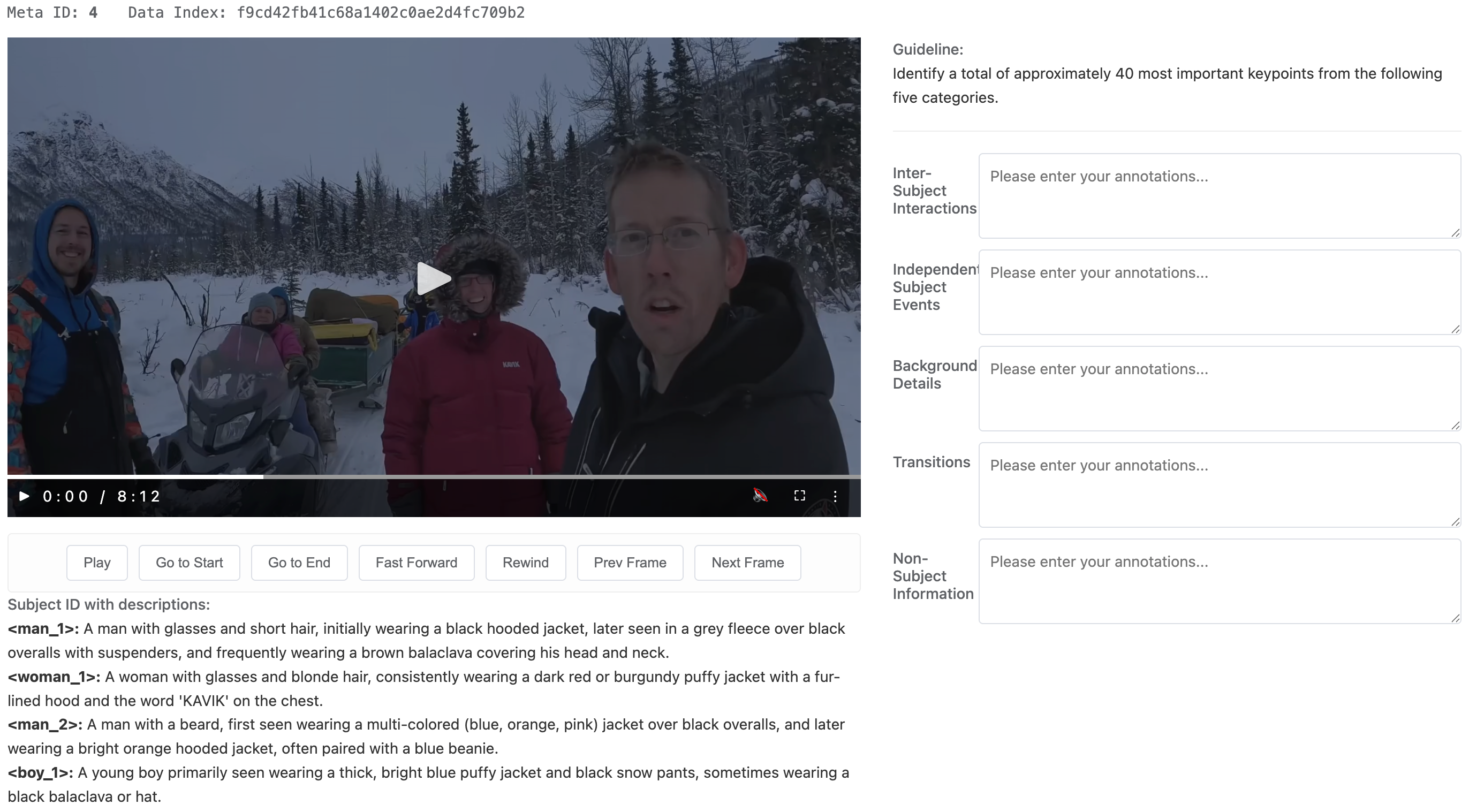}
    \caption{Screenshot of the annotation system interface.}
    \label{fig:label}
\end{figure*}

In the subsequent annotation stage, we recruit twenty experienced multilingual annotators through the same crowdsourcing platform to participate in the labeling process. To illustrate the annotation workflow, we provide a screenshot of the annotation interface in Figure~\ref{fig:label}, which demonstrates how annotators interact with the system and complete annotation tasks.

To ensure the quality and reliability of both the video collection and annotation, annotators are compensated based on the time spent rather than the number of samples completed, thereby reducing incentives for rushed or superficial work. Annotators are paid at a rate of USD~10 per hour, which is highly competitive relative to prevailing industry standards for comparable tasks.

\begin{table}[t]
    \centering
    \resizebox{\linewidth}{!}{
    \begin{tabular}{ll}
    \toprule
    \textbf{Major Category} & \textbf{Subcategories} \\
    \midrule
    Relationship & \makecell[l]{Friendship \& Companionship; Romantic\\Love; Professional Ties; Mentorship;\\Community Life}\\
    \midrule
    Youth & \makecell[l]{Campus Life; Coming-of-Age; Ambition\\\& Dreams; Transition to Adulthood}\\
    \midrule
    Entertainment & \makecell[l]{Sketch Comedy; Variety \& Reality Shows;\\Lighthearted Drama; Sports Competition;\\Musical \& Dance Performances}\\
    \midrule
    History & \makecell[l]{Culture Heritage; Politics Affairs;\\Military Conflict; Society Evolution;\\Historical Biography; Memory}\\
    \midrule
    Family & \makecell[l]{Family Bonding; Mutual Support;\\Family Conflict; Everyday Leisure;\\Parenting \& Education}\\
    \midrule
    Lifestyle & \makecell[l]{Urban Living; Rural Living; Nature \&\\Outdoors; Home \& Settlement; Travel\\\& Exploration}\\
    \midrule
    Fantasy & \makecell[l]{Adventure \& Exploration; Science Fiction;\\Supernatural Themes}\\
    \midrule
    Mystery & \makecell[l]{Deduction \& Detective; Crime Narratives;\\Psychological Games}\\
    \bottomrule
    \end{tabular}
    }
    \caption{Detailed video categories of CapRiCorn-1K.}
    \label{tab:full_category}
\end{table}

\section{Further Analysis}
\label{app:further_analysis}
In this section, we provide additional analyses from four distinct perspectives: (1) the relationship between caption length and our evaluation metrics; (2) the trade-off between the number of input frames and input resolution under a fixed context-window budget; (3) the impact of the maximum input frame count under a fixed resolution; and (4) the impact of the maximum input resolution under a fixed frame count. The corresponding results are presented in Figure~\ref{fig:further_analysis}.

\subsection{Analysis on Caption Length}
Some caption evaluation benchmarks can obtain artificially higher scores simply by encouraging models to generate longer captions, which fails to faithfully reflect the actual quality of the captions. To verify that our evaluation metrics are not strongly correlated with caption length, we select several representative captioning models and analyze the correlation between their performance on CapRiCorn-1K and their average caption lengths, as illustrated in Figure~\ref{fig:further_analysis}a. The results demonstrate that the evaluation metrics of CapRiCorn-1K are not directly associated with caption length. Specifically, the Pearson correlation coefficients between caption length and Acc, Cov, and Ref are 0.525, 0.561, and 0.429, respectively.

\subsection{Analysis on Frame Count and Resolution}
In the main experiments, our evaluation setting prioritizes relatively high spatial resolution (typically 512 $\times$ 512) while determining the maximum number of frames according to the context-window limit of each model. To investigate the trade-off between frame count and resolution under a fixed context-window budget, we take Qwen3-Omni-Captioner as a case study. Specifically, we constrain the total number of visual tokens to approximately 25K while varying the maximum frame count and resolution for analysis. The results are presented in Figure~\ref{fig:further_analysis}b.

The results show that excessively high resolution (which consequently leads to an insufficient number of frames), as well as excessively large frame counts (which consequently require overly low resolution), both lead to performance degradation. Therefore, maintaining a sufficiently high resolution and then increasing the number of frames within the context window budget is more beneficial for captioning performance.

\subsection{Analysis on Frame Count Only}
\begin{figure}[t]
    \centering
    \includegraphics[width=\linewidth]{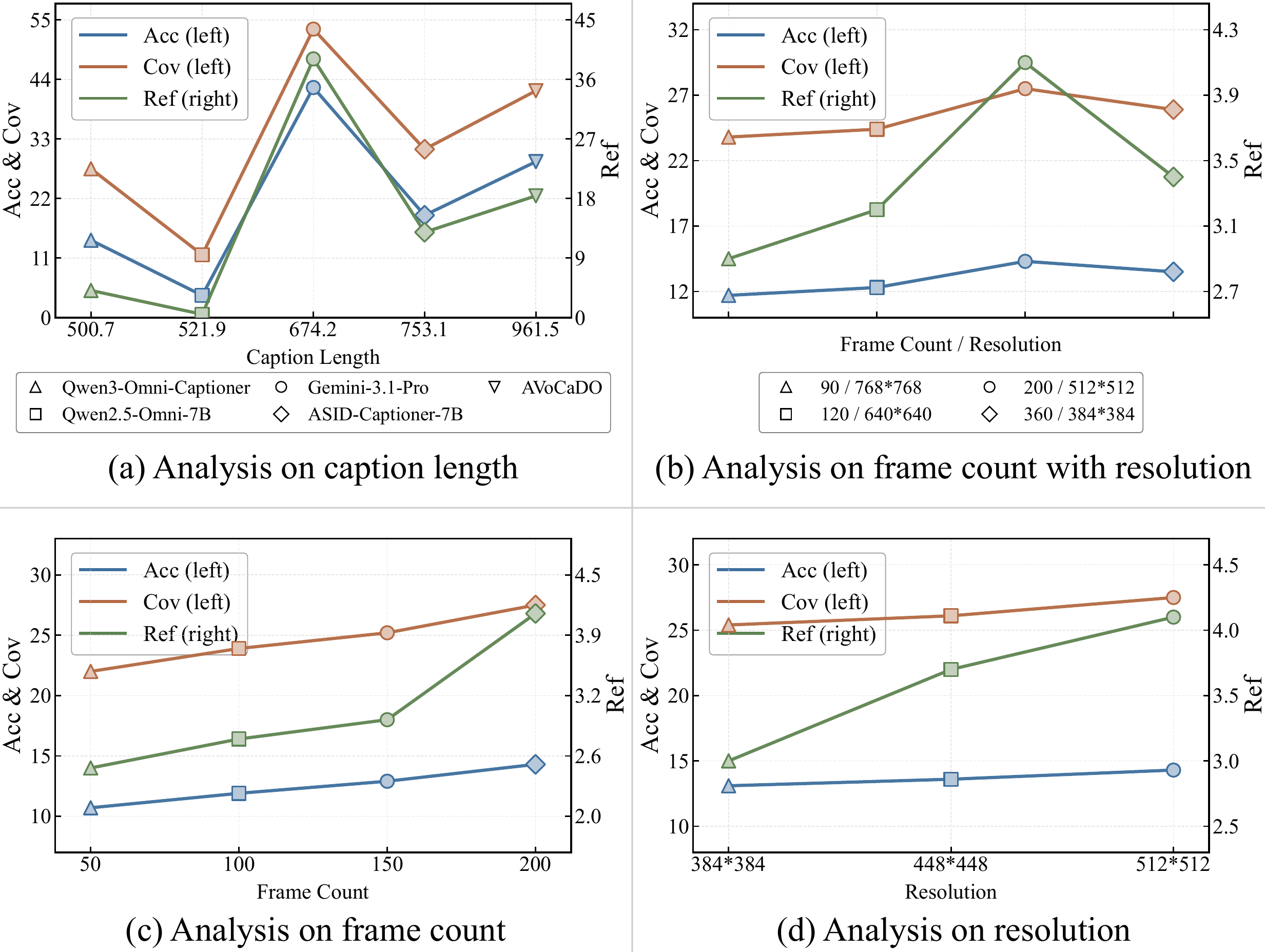}
    \caption{Further analysis of captioning performance with respect to (a) caption length; (b) the trade-off between frame count and resolution under a fixed context-window budget; (c) frame count under a fixed resolution; and (d) resolution under a fixed frame count.}
    \label{fig:further_analysis}
\end{figure}

To independently evaluate the effect of the maximum number of input frames, we conduct an ablation study on the maximum input frame count while fixing the input resolution of Qwen3-Omni-Captioner to 512 $\times$ 512, as shown in Figure~\ref{fig:further_analysis}c. The results indicate that, within the limitation of the maximum context window, captioning performance consistently improves as the maximum number of input frames increases.

\subsection{Analysis on Resolution Only}
To independently evaluate the effect of the maximum input resolution, we conduct an ablation study on the input resolution while fixing the number of input frames of Qwen3-Omni-Captioner to 200, as shown in Figure~\ref{fig:further_analysis}d. The results demonstrate that, within the limitation of the maximum context window, the captioning performance consistently improves as the maximum input resolution increases.

\subsection{Error Analysis}
Through qualitative examination of failure cases, we identify three representative scenarios in CapRiCorn-1K that remain particularly challenging for current models. 

\begin{figure*}[t]
    \centering
    \includegraphics[width=\linewidth]{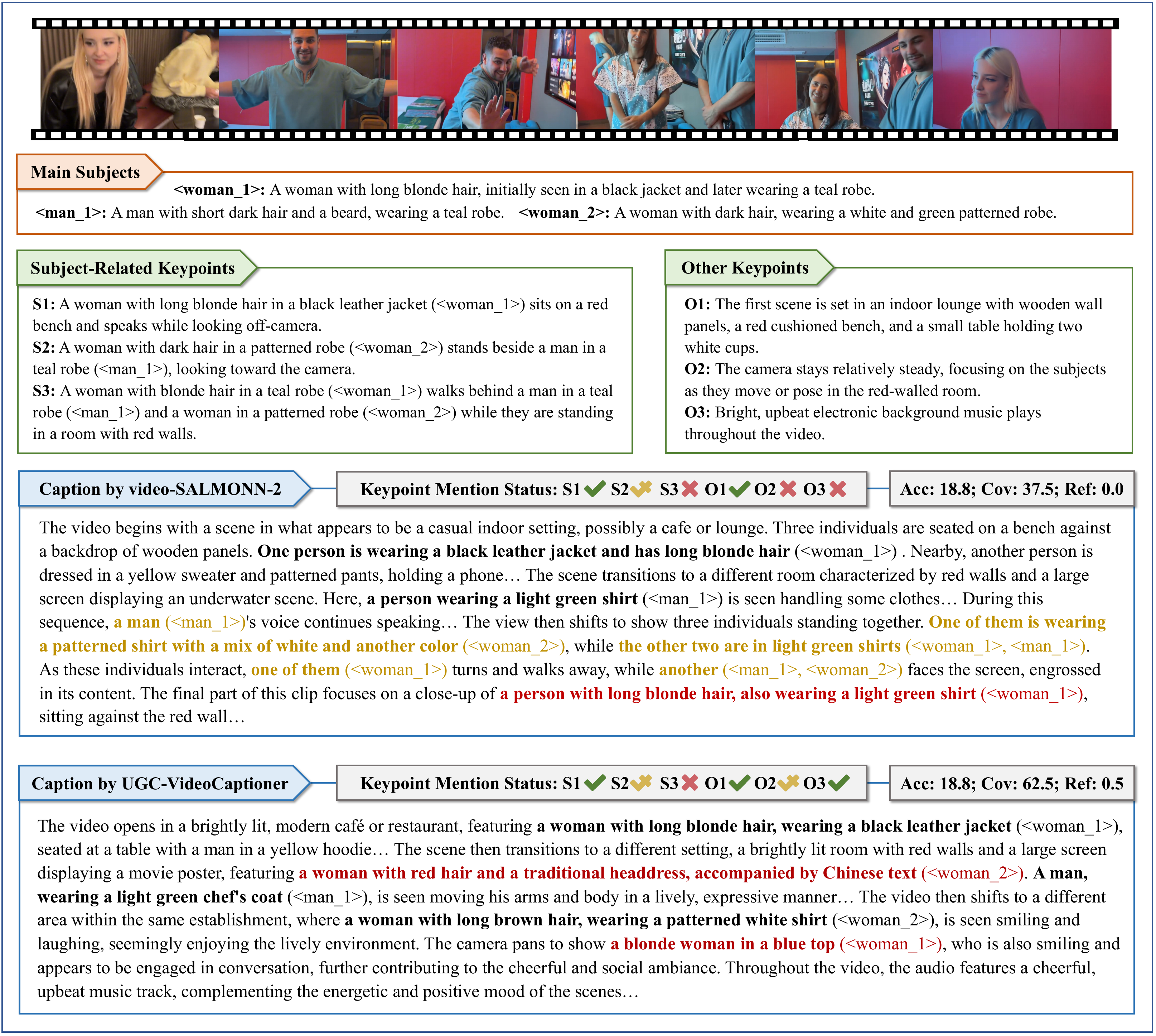}
    \caption{Error analysis for clothing-change scenarios. 
        Keypoints marked with \raisebox{-0.2em}{\includegraphics[height=1em]{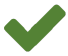}}, \raisebox{-0.2em}{\includegraphics[height=1.15em]{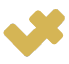}}, and \raisebox{-0.2em}{\includegraphics[height=1em]{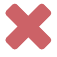}} are ``correctly mentioned'', ``partially mentioned or containing errors'', and ``not mentioned'', respectively.
        Subject descriptions are highlighted in \textbf{bold} with the ground-truth subject-ID shown in parentheses immediately afterward. Different colors representing 
        \textcolor[HTML]{548135}{\textbf{consistent and correct}}, 
        \textcolor[HTML]{bd8f03}{\textbf{ambiguous}}, and 
        \textcolor[HTML]{ae0405}{\textbf{inconsistent or incorrect}} subject references, respectively.}
    \label{fig:error_cloth_change}
\end{figure*}

\begin{figure*}[t]
    \centering
    \includegraphics[width=\linewidth]{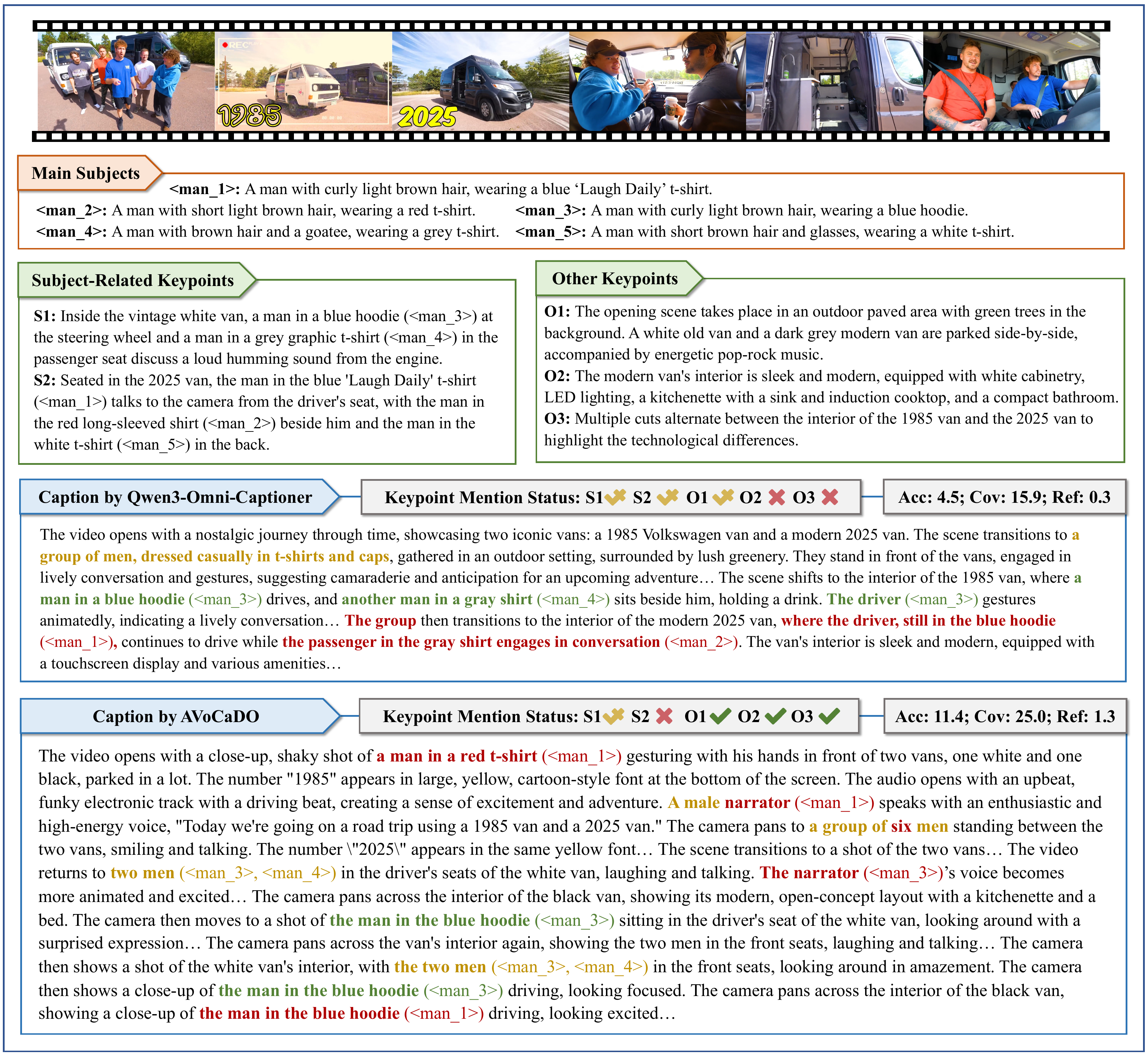}
    \caption{Error analysis in multi-subject scenarios. 
        Keypoints marked with \raisebox{-0.2em}{\includegraphics[height=1em]{figs/correct_icon.pdf}}, \raisebox{-0.2em}{\includegraphics[height=1.15em]{figs/partial_icon.pdf}}, and \raisebox{-0.2em}{\includegraphics[height=1em]{figs/wrong_icon.pdf}} are ``correctly mentioned'', ``partially mentioned or containing errors'', and ``not mentioned'', respectively.
        Subject descriptions are highlighted in \textbf{bold} with the ground-truth subject-ID shown in parentheses immediately afterward. Different colors representing 
        \textcolor[HTML]{548135}{\textbf{consistent and correct}}, 
        \textcolor[HTML]{bd8f03}{\textbf{ambiguous}}, and 
        \textcolor[HTML]{ae0405}{\textbf{inconsistent or incorrect}} subject references, respectively.}
    \label{fig:error_multiperson}
\end{figure*}

\begin{figure*}[t]
    \centering
    \includegraphics[width=\linewidth]{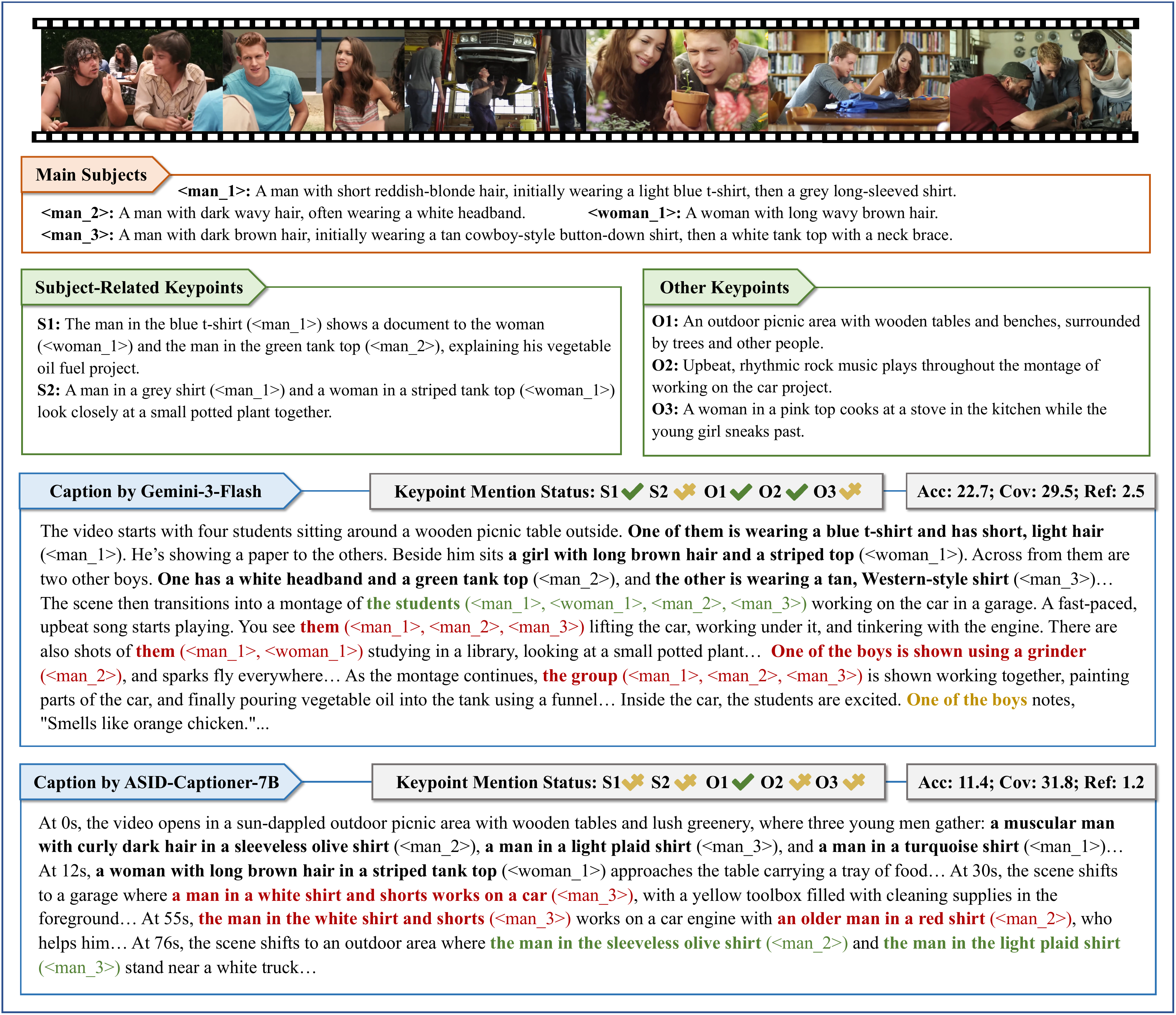}
    \caption{Error analysis in multi-scene scenarios. 
        Keypoints marked with \raisebox{-0.2em}{\includegraphics[height=1em]{figs/correct_icon.pdf}}, \raisebox{-0.2em}{\includegraphics[height=1.15em]{figs/partial_icon.pdf}}, and \raisebox{-0.2em}{\includegraphics[height=1em]{figs/wrong_icon.pdf}} are ``correctly mentioned'', ``partially mentioned or containing errors'', and ``not mentioned'', respectively.
        Subject descriptions are highlighted in \textbf{bold} with the ground-truth subject-ID shown in parentheses immediately afterward. Different colors representing 
        \textcolor[HTML]{548135}{\textbf{consistent and correct}}, 
        \textcolor[HTML]{bd8f03}{\textbf{ambiguous}}, and 
        \textcolor[HTML]{ae0405}{\textbf{inconsistent or incorrect}} subject references, respectively.}
    \label{fig:error_multiscene}
\end{figure*}

\begin{itemize}[leftmargin=*]
\item \textbf{Clothing Changes} (Figure~\ref{fig:error_cloth_change}). In such scenarios, the model should not only precisely describe different outfits worn by the same subject, but also explicitly articulate the clothing transitions between them to maintain consistent subject tracking throughout the caption. However, reliably capturing and narrating such wardrobe changes for a single subject remains a persistent challenge for existing models.

\item \textbf{Multiple Subjects} (Figure~\ref{fig:error_multiperson}). As the number of subjects increases, models tend to confuse referential relationships among different subjects, particularly when multiple subjects share similar visual appearances or attributes.

\item \textbf{Multiple Scenes} (Figure~\ref{fig:error_multiscene}). Frequent scene transitions prevent the model from distinguishing subjects based on relatively stable positional cues within a single scene, thereby increasing the difficulty of maintaining consistent subject references. As a result, models often resort to generating ambiguous references or producing incorrect referential associations.
\end{itemize}

\section{Implementation Details}
\label{app:imple_detail}
Our evaluation adheres to the official protocols of each model by default. When such protocols are unavailable, given the substantial length of the videos, we uniformly sample frames up to the maximum context window supported by the model while preserving sufficient frame resolution. In this section, we provide the detailed evaluation settings for all models, which is also summarized in Table~\ref{tab:imple_detail}.

\begin{table}[t]
    \centering
    \resizebox{\linewidth}{!}{
    \begin{tabular}{l|ccc}
    \toprule
    \textbf{Model Class} & \textbf{Max Pixels per Frame} & \textbf{FPS} & \textbf{Max Frames} \\
    \midrule
    \multicolumn{4}{l}{\textcolor{gray!100}{\textit{Audiovisual Models}}} \\
    Qwen2.5-Omni & 200,704 (448$\times$448) & 2 & 200 \\
    Qwen3-Omni & 262,144 (512$\times$512) & 2 & 200 \\
    video-SALMONN-2 & 147,456 (384$\times$384) & 1 & 110 \\
    video-SALMONN-2+ & 61,250 (\textasciitilde248$\times$248) & 10 & 768 \\
    OmniVinci & original & 2 & 128 \\
    ARC-Qwen-Video & 153,664 (392$\times$392) & 1 & 300 \\
    \midrule
    \multicolumn{4}{l}{\textcolor{gray!100}{\textit{Vision-Only Models}}} \\
    Qwen3-VL & 262,144 (512$\times$512) & 2 & 768 \\
    Qwen3.5 & 262,144 (512$\times$512) & 2 & 768 \\
    Qwen3.6 & 262,144 (512$\times$512) & 2 & 768 \\
    InternVL3.5 & 200,704 (448$\times$448) & 2 & 100 \\
    MiMo-VL & 100,352 (224$\times$224) & 2 & 200 \\
    Tarsier2 & 460,800 (640$\times$720) & - & 256 \\
    \bottomrule
    \end{tabular}
    }
    \caption{Implementation details of the evaluation settings. Frames are initially sampled at the target FPS. If the resulting frame count exceeds the Max Frames limit, uniform sampling is applied to satisfy the constraint.}
    \label{tab:imple_detail}
\end{table}

For Qwen2.5-Omni-style models, including Qwen2.5-Omni, UGC-VideoCaptioner, AVoCaDO, DiaDem, and ASID-Captioner, we set the maximum number of vision tokens per frame to 256, corresponding to 200,704 pixels per frame (i.e., $256 \times 14 \times 14 \times 2 \times 2$). The frame rate is fixed at 2 FPS. Given a maximum context window of 32K tokens, we set the maximum number of frames to 200, resulting in a maximum visual token length of 25,600 (i.e., $256 \times 200 / 2$ after accounting for temporal aggregation). For Qwen3-Omni-style models, although the maximum context window is 64K, their technical report indicates that training is conducted only up to 32K context length. We therefore adopt the 32K configuration for evaluation to ensure a consistent and comparable setting. All other audiovisual models are evaluated using their default official configurations without modification.

For Qwen3-VL-style models, including Qwen3-VL, Qwen3.5, and Qwen3.6, we set the maximum number of vision tokens per frame to 256, corresponding to 262,144 pixels per frame (i.e., $256 \times 16 \times 16 \times 2 \times 2$), with 2 FPS and a maximum of 768 frames following the recommended long-video setting. For InternVL3.5, we use the official 448 $\times$ 448 input resolution, corresponding to 200,704 pixels per frame and set the max number of frames to 200 under the 32K context budget to better support long-video evaluation. For MiMo-VL, we set the max pixels to 100,352 (i.e., $128 \times 14 \times 14 \times 2 \times 2$), and use 2 FPS with a maximum of 200 frames under its 16K context constraint. For Tarsier2, we keep its default maximum pixel budget of 460,800 pixels per frame and increase the frame number from the default 16 to the supported maximum of 256. 

All models evaluated in this work are strictly limited to academic research purposes and comply with their respective official licenses. For all statistical analyses involving Pearson correlation coefficients, we use SciPy version 1.14.1.

\section{Prompt Details}
Figures~\ref{fig:prompt_sbj_keypoints} and~\ref{fig:prompt_nonsbj_keypoints} illustrate the initial evaluation step of CapRiCorn-1K, which involves determining the mention status of each keypoint in the caption, thereby enabling an assessment of overall caption quality. For the subject-related keypoints that are correctly or partially mentioned, the corresponding localized subject descriptions are simultaneously extracted from the caption. Descriptions associated with the same ground-truth subjects are then clustered within the caption context to evaluate referential consistency, using the prompt in Figure~\ref{fig:prompt_clustering}.

Figures~\ref{fig:av_prompt_list} and~\ref{fig:v_prompt_list} present the prompt lists used to generate captions for audiovisual video captioning models and vision-only video captioning models, respectively. These prompts are randomly sampled to assess both general captioning capabilities and the ability to maintain subject referential consistency within the generated captions.

\begin{figure*}
\begin{tcolorbox}[colback=white!95!gray, colframe=gray!50!black, rounded corners, title={Joint assessment of mention status and subject-description extraction for subject-related keypoints}]
You will be given a video caption and a specific event involving one or more subjects (predefined and enclosed in <>; do NOT add any subjects that you think should be included). Your task is to determine how this event is represented in the caption by assigning it to one of the following categories:\\
- "correctly mentioned": the event is accurately described in the caption, possibly with only minor omissions or negligible errors.\\
- "mentioned but with errors": the event is mentioned in the caption, but contains substantial inaccuracies, distortions, or misleading details.\\
- "not mentioned": the event is not described in the caption at all.\\

If the classification is "correctly mentioned" or "mentioned but with errors":\\
- From the **local caption segment** at the moment the event occurs, identify the subject ID enclosed in <> within the event, and extract the corresponding subject description from the **local caption segment**. Only when the local subject description is overly vague (e.g., “his”, “her”, “their”, “it”, “the man”) is it allowed to use the global context to obtain a more specific subject description.\\
- The extracted content must be strictly from the local descriptions present in the caption text, NOT copied or inferred from the provided subject description in the event.\\
- Each subject’s local description should be concise while still containing enough identifying information.\\

If the classification is "not mentioned":\\
- Set the subject description value to null.\\\\

Video caption:\\
\{\}\\

Event:\\
\{\}

Subject ID list:\\
\{\}\\

Output format:\\
\textasciigrave\textasciigrave\textasciigrave json\\
\{\{\\
    "event\_type": "xx",  // One of ["correctly mentioned", "mentioned but with errors", "not mentioned"]\\
    "reason": "xx",  // Brief justification for event\_type; no double quotes inside\\
    "subject\_description\_in\_caption": \{\{\\
        "<sbj\_id\_1>": xx,\\
        "<sbj\_id\_2>": xx, // if exists\\
        ...\\
    \}\}  // **Brief subject descriptions dict (rather than event description)** summarized from caption or null (only when the event\_type is "not mentioned"). **Do not use any pronouns (e.g., his, her, their, it)**; instead, replace them with their corresponding referents identified from the caption.\\
\}\}\\
\textasciigrave\textasciigrave\textasciigrave
\end{tcolorbox}
\caption{Prompts to jointly evaluate the mention status of subject-related keypoints and extract subject descriptions for the mentioned keypoints.}
\label{fig:prompt_sbj_keypoints}
\end{figure*}

\begin{figure*}
\begin{tcolorbox}[colback=white!95!gray, colframe=gray!50!black, rounded corners, title={Prompts to evaluate the mention status of other keypoints not related to the subject}]
You will be given a video caption and a specific event. Your task is to determine how this event is represented in the caption by assigning it to one of the following categories:\\
- "correctly mentioned": the event is accurately described in the caption, possibly with only minor omissions or negligible errors.\\
- "mentioned but with errors": the event is mentioned in the caption, but contains substantial inaccuracies, distortions, or misleading details.\\
- "not mentioned": the event is not described in the caption at all.\\

Video caption:\\
\{\}\\

Event:\\
\{\}\\

Output format:\\
\textasciigrave\textasciigrave\textasciigrave json\\
\{\{\\
    "event\_type": "xx",  // One of ["correctly mentioned", "mentioned but with errors", "not mentioned"]\\
    "reason": "xx",  // Brief justification for event\_type; no double quotes inside\\
\}\}\\
\textasciigrave\textasciigrave\textasciigrave
\end{tcolorbox}
\caption{Prompts to evaluate the mention status of other keypoints not related to the subjects.}
\label{fig:prompt_nonsbj_keypoints}
\end{figure*}

\begin{figure*}
\begin{tcolorbox}[colback=white!95!gray, colframe=gray!50!black, rounded corners, title={Prompts for clustering descriptions of the same ground-truth subject}]
You will be given a list of subject descriptions and a video caption. Your task is to group these descriptions into clusters, where each cluster contains descriptions that refer to the same real-world subject, based on both the descriptions and the caption context.\\

Descriptions should be grouped together only if they satisfy at least **one of** the following conditions:\\
1. They share sufficiently specific matching **appearance attributes**, without considering actions.\\
2. They contain the same subject name. In this case, attribute differences must be ignored, and **all descriptions with the same subject name must always be grouped into a single cluster**.\\
3. Based on the video caption, it can be reasonably and clearly inferred that the descriptions refer to the same subject.\\

Guidelines:\\
- Note that identical descriptions do not necessarily refer to the same subject.\\
    - For example, multiple generic references such as “a girl” should be treated as distinct subjects, because the only feature "girl" is too vague to determine that they refer to the same entity, unless the caption clearly implies they refer to the same entity (e.g., there is only one girl in the caption).\\
    - Similarly, ambiguous descriptions like "one of xxx" or "two other xxx", which lack distinguishing details, should be treated as referring to different subjects (i.e., if the phrase "one of xxx" appears four times, it should be classified as four **distinct** categories), unless the caption provides sufficient evidence to identify them as the same entity.\\

- Conversely, non-identical descriptions may still refer to the same subject, as long as they convey consistently matching attributes, share identical subject names, or can be reasonably and clearly inferred from the caption.\\
    - For example, descriptions with similar attributes such as “a boy with a light-grey shirt” and “the boy in a grey shirt” should be grouped into the same category.\\
    - Similarly, descriptions that include the same subject name, such as “James”, “James in a white shirt” and “James, dressed in a dark suit jacket”, should also be grouped into one cluster, even if their outfits differ, because they contain the same subject name.\\

- Ensure that every subject description is assigned to exactly one cluster.\\

List of subject descriptions:\\
\{\}\\

Video Caption:\\
\{\}\\

Output format:\\
\textasciigrave\textasciigrave\textasciigrave json\\
\{\{\\
    "category\_1": ["original subject description 1", "original subject description 2", ...],\\
    "category\_2": ["original subject description 3"], // optional\\
    ...\\
\}\}\\
\textasciigrave\textasciigrave\textasciigrave
\end{tcolorbox}
\caption{Prompts for clustering descriptions of the same ground-truth subject.}
\label{fig:prompt_clustering}
\end{figure*}

\begin{figure*}
\begin{tcolorbox}[colback=white!95!gray, colframe=gray!50!black, rounded corners, title={List of prompts used to evaluate audiovisual video captioning models}]
1. Provide a comprehensive description of all the content in the video, leaving out no details. Be sure to include as much of the audio information as possible, and ensure that your descriptions of the audio and video are closely aligned. Ensure coherence in the description of the same subject throughout.\\
2. Thoroughly describe everything in the video, capturing every detail. Include as much information from the audio as possible, and ensure that the descriptions of both audio and video are well-coordinated. Ensure coherence in the description of the same subject throughout.\\
3. Please describe all the information in the video without sparing every detail in it. As you describe, you should also describe as much of the information in the audio as possible, and pay attention to the synchronization between the audio and video descriptions. Ensure coherence in the description of the same subject throughout.\\
4. Offer a detailed description of the video, making sure to include every detail. Also, incorporate as much information from the audio as you can, and ensure that your descriptions of the audio and video are in sync. Ensure coherence in the description of the same subject throughout.\\
5. Describe every aspect of the video in full detail, covering all the information it contains. Additionally, include as much of the audio content as you can, and make sure your descriptions of the audio and video are synchronized. Ensure coherence in the description of the same subject throughout.\\
6. Please provide a thorough description of all the content in the video, including every detail. As you describe, ensure that you also cover as much information from the audio as possible, and be mindful of the synchronization between the audio and video as you do so. Ensure coherence in the description of the same subject throughout.\\
7. Give a detailed account of everything in the video, capturing all the specifics. While doing so, also include as much information from the audio as possible, ensuring that the descriptions of audio and video are well-synchronized. Ensure coherence in the description of the same subject throughout.
\end{tcolorbox}
\caption{List of prompts used to evaluate the audiovisual video captioning models. During evaluation, prompts are randomly sampled from this list.}
\label{fig:av_prompt_list}
\end{figure*}

\begin{figure*}
\begin{tcolorbox}[colback=white!95!gray, colframe=gray!50!black, rounded corners, title={List of prompts used to evaluate vision-only video captioning models}]
1. Provide a comprehensive description of all visible content in the video, leaving out no important visual details. Describe the subjects, actions, scenes, objects, camera changes, and temporal progression as clearly as possible. Ensure coherence in the description of the same subject throughout.\\
2. Thoroughly describe everything that can be observed in the video. Include detailed information about the people or subjects, their appearances, actions, interactions, background, scene transitions, and changes over time. Ensure coherence in the description of the same subject throughout.\\
3. Please describe all visual information in the video in detail. Focus on the subjects, their actions, spatial relationships, environment, objects, scene changes, and the overall temporal sequence of events. Ensure coherence in the description of the same subject throughout.\\
4. Offer a detailed visual description of the video, making sure to cover important subjects, actions, interactions, background details, object appearances, camera movements, and scene transitions. Ensure coherence in the description of the same subject throughout.\\
5. Describe every visible aspect of the video in full detail. Pay attention to the identities and appearances of recurring subjects, their actions, interactions, locations, and how the scene evolves over time. Ensure coherence in the description of the same subject throughout.\\
6. Please provide a thorough visual description of the video, including all important details. Describe what happens from beginning to end, and maintain consistent references to the same subjects throughout the description. Ensure coherence in the description of the same subject throughout.\\
7. Give a detailed account of the visual content in the video, capturing the subjects, objects, actions, backgrounds, scene transitions, and temporal order of events. Ensure that recurring subjects are described coherently and consistently. Ensure coherence in the description of the same subject throughout.
\end{tcolorbox}
\caption{List of prompts used to evaluate the vision-only video captioning models. During evaluation, prompts are randomly sampled from this list.}
\label{fig:v_prompt_list}
\end{figure*}

\end{document}